\documentclass[runningheads]{llncs}

\usepackage[T1]{fontenc}
\usepackage[utf8]{inputenc}

\usepackage{microtype}
\usepackage{inconsolata}

\usepackage{enumitem}

\usepackage{xcolor}

\usepackage{graphicx}
\graphicspath{ {./img/} }

\usepackage{mathtools}
\usepackage{bm}

\usepackage{booktabs}
\usepackage{tabularx}
\usepackage{multirow}

\usepackage[frozencache, cachedir=minted-cache]{minted}
\makeatletter
    \renewcommand{\fnum@listing}{\textbf{Listing~\thelisting.}}
    \renewcommand\floatc@plain[2]{{\small #1 #2\par}}
\makeatother

\usepackage[bookmarksdepth=2]{hyperref}
\usepackage{cleveref}
\usepackage{bookmark}
\usepackage[all]{hypcap}
\usepackage{cite}

\usepackage{pifont}
\usepackage[fixed]{fontawesome5}

\newcommand{\eg}{e.g., }
\newcommand{\ie}{i.e., }
\newcommand{\cf}{cf. }
\newcommand{\etal}{et al. }

\newcounter{descctr}
\newcommand{\desclabel}[1]{\refstepcounter{descctr}\label{#1}}
\newcommand{\descref}[1]{RQ\hyperref[#1]{#1}}

\begin{document}

\title{A Comparative Study on Affective Cues in Text Embeddings Across Psychological Emotion Theories}

\titlerunning{Affective Cues in Text Embeddings Across Emotion Theories}

\author{
    Fabio Ciani\inst{1,}\thanks{Equal contribution.} \and
    Harald Schweiger\inst{2,}\textsuperscript{\thefootnote} \and\\
    Emilia Parada-Cabaleiro\inst{3} \and
    Markus Schedl\inst{2,4}
}

\authorrunning{Ciani et al.}

\institute{
    Department of Electronics, Information, and Bioengineering,\\Polytechnic University of Milan\\
    \email{fabio1.ciani@mail.polimi.it} \and
    Multimedia Mining and Search Group, Institute of Computational Perception, Johannes Kepler University Linz\\
    \email{\{harald.schweiger,markus.schedl\}@jku.at} \and
    Department of Music Pedagogy, Nuremberg University of Music\\
    \email{emiliaparada.cabaleiro@hfm-nuernberg.de} \and
    Human-centered AI Group, AI Lab, Linz Institute of Technology
}

\maketitle

\setcounter{footnote}{0}

\begin{abstract}
    Text encoders are known for their utility in natural language processing, as they are able to efficiently compress inputs into dense vectors while preserving semantics.
These models have been applied to affective computing, in particular to help with solving sentiment analysis and emotion recognition tasks.
Nevertheless, it remains unclear to what extent the latent representations produced by modern text encoders capture well-defined psychological theories of affect.
In this work, we investigate the affective capabilities of twelve recently released text encoders by probing their generated embeddings as input features for solving regression and classification tasks across three established emotion frameworks, using both word- and sentence-level data.
Additionally, we apply a semantic data-leakage prevention technique to improve robustness in word-level evaluations.
Our main findings show that the latent manifolds of the latest instruction-aware open-weight encoders enclose an equal or even a larger amount of affective information in comparison with proprietary counterparts when evaluated at word level.
In contrast, embeddings of task-tuned and proprietary encoders reach the highest scores on sentence-level affective classification.
Furthermore, a qualitative analysis of latent representations and their encoded affective cues is provided.

\end{abstract}

\section{Introduction}

The advent of text encoders has transformed the way of converting textual content into numerical representations, known as embeddings, which now serve as core components for a variety of tasks, \eg semantic similarity, retrieval, and reranking \cite{Chen2024}.
They have also proven to be valuable for text-based sentiment analysis, \eg in emotion classification \cite{Demszky2020} or valence-arousal regression \cite{Ito2022}.
To solve these tasks, models such as \texttt{BERT} \cite{Devlin2019,Reimers2019} have been tested under various configurations, mostly involving end-to-end fine-tuning.
Since then, the performance of text encoders has been progressively improved by different techniques, including instruction-based queries and advanced training schemes \cite{Zhao2025}.

Despite relevant prior works, the usage of text encoders as zero-shot feature extractors for emotion recognition tasks remains underexplored, especially with respect to the latest state-of-the-art and instruction-aware models.
Moreover, it is uncertain whether the latent manifolds induced by these models enclose established emotion frameworks from psychology through a sufficient encoding of affective information.

To fill this gap, our analysis compares recently released text encoders, without prior fine-tuning, across different emotion theories, \ie the Pleasure-Arousal-Dominance model by Mehrabian and Russell \cite{Mehrabian1974}, the model of emotions by Plutchik \cite{Plutchik1980}, and the “big six” by Ekman \cite{Ekman1971}.
We use two structured lexica and one sentence-level dataset, \ie NRC-VAD \cite{Mohammad2025}, NRC-EIL \cite{Mohammad2018a}, and GoEmotions \cite{Demszky2020}, which respectively correspond to each of the emotion theories, together with a novel technique to limit leakage between splits.
Accordingly, embeddings are computed and frozen to be exploited as input features to four downstream predictors and subsequently evaluated with respect to their affective cues. The quantitative results, together with a qualitative visual analysis, address the following research questions.

\begin{description}[labelwidth=\widthof{\textbf{RQ0}}, leftmargin=!]
    \item[RQ1]\desclabel{1} \textit{To what extent do latent manifolds from text encoders enclose emotional cues?}
    \item[RQ2]\desclabel{2} \textit{Are instruction-aware text encoders superior to task-tuned models or ones without explicit prompt support for generating optimized embeddings?}
    \item[RQ3]\desclabel{3} \textit{Are proprietary models better than open-weight ones?}
    \item[RQ4]\desclabel{4} \textit{Does model performance vary depending on the chosen emotion framework and downstream predictor?}
\end{description}

The paper proceeds as follows.
In \Cref{sec:literature}, a background on emotion theories and affective language processing is presented.
\Cref{sec:methods} describes the experimental setup, including datasets, encoders, and predictors.
Finally, \Cref{sec:results} reports the results, while \Cref{sec:outro} concludes the article.\footnote{Our code can be publicly accessed at the repository at the following hyperlink: \url{https://github.com/hcai-mms/affective_embeddings}.}

\section{Related works}\label{sec:literature}

\subsection{Emotion theories}

Consolidated research in psychology has led to a variety of frameworks explaining human emotions from both taxonomic and perceptual standpoints.
One of the first categorical models was proposed by Ekman, in which the so called “big six” basic emotions (\emph{anger}, \emph{disgust}, \emph{fear}, \emph{happiness}, \emph{sadness}, and \emph{surprise}) were identified from facial expressions and considered biologically encoded and cross-cultural \cite{Ekman1971}.

Another notable theory is the circumplex model by Russell, which represents affects in a two-dimensional space, where the horizontal and vertical axes respectively measure valence and arousal.
While valence captures the pleasure ranging from negative to positive, arousal reflects the energy spanning from low to high \cite{Russell1980}.
This was preceded by the PAD (Pleasure-Arousal-Dominance) emotion model by Mehrabian, where a third bipolar dimension to quantify the evoked control or submissiveness was also included \cite{Mehrabian1974}.

An additional framework, also relevant to affective computing, is the one by Plutchik, who designed a hybrid categorical-dimensional model with a resemblance to Russell's theory, in which spatial proximity links to affect similarity.
Eight primary emotions (\emph{joy}, \emph{trust}, \emph{fear}, \emph{surprise}, \emph{sadness}, \emph{disgust}, \emph{anger}, and \emph{anticipation}) are arranged in concentric circles corresponding to different levels of intensity, \ie as a cone subdivided in sectors, with the possibility to mix adjacent and opposite emotions to form combined mood dyads \cite{Plutchik1980}.

\subsection{Affective language processing}

Early techniques to extract emotions from textual content and drawing from the distributional hypothesis in linguistics \cite{Harris1954,Firth1957} included latent semantic analysis \cite{Bellegarda2013}, a matrix factorization procedure to learn compressed representations, that later evolved into popular self-supervised word embeddings \cite{Mikolov2013a,Mikolov2013b,Pennington2014,Bojanowski2017}.

Word vectors were retrained from scratch incorporating supervised affective contexts into the objective function \cite{Tang2016}.
Faruqui \etal \cite{Faruqui2015} and Mrksic \etal \cite{Mrksic2016} devised a method for adjusting pre-trained word embeddings with respect to lexical relationships and constraints, which was adopted by Yu \etal \cite{Yu2017} and Seyeditabari \etal \cite{Seyeditabari2019} on affective datasets to mitigate reported issues in vector similarity and arithmetic associated with general-purpose distributional embeddings \cite{Seyeditabari2017}.
Notable extensions built upon the Transformer architecture have been presented both at word level \cite{Chochlakis2023a,Chochlakis2023b} and at sentence level \cite{Shah2023}.
The attention mechanism has also been adapted to enrich learnt representations by combining vectors and data from a knowledge base \cite{Suresh2021}.

More broadly, it has been questioned whether language models (LMs) can effectively understand the multifaceted nature of emotions.
Lee \etal \cite{Lee2025a} isolated low-level subcomponents focused on handling patterns deriving from specific affects, while Reichman \etal \cite{Reichman2025} continued the analysis underlining the presence of a complex redundancy scheme implemented by sets of specialized neurons and connections within the neural architecture.
At a higher level, the neuropsychology of LMs has been studied by assessing whether their internal representations can be refined to align with established emotion theories \cite{Lee2023}.
It has also been observed that larger foundational models tend to exhibit emotional intelligence more accurately than smaller counterparts \cite{Wang2023} and build increasingly detailed hierarchical taxonomies to organize emotions \cite{Zhao2024}.

Lastly, connecting the affective expression in an emotion model to the definition in another theoretical framework has been achieved through annotated textual content, \ie data with a series of assigned labels or numerical quantities, either directly bridging the categorical and dimensional families \cite{Park2021} or upon learning an agnostic intermediate representation space for conversions \cite{Buechel2021}.

\section{Methodology}\label{sec:methods}

To test the ability of text encoders to capture emotional cues, we performed evaluations on three corpora grounded in different emotion frameworks from psychology (\cf \Cref{sec:data}).
We favored structured lexica, \ie NRC-VAD \cite{Mohammad2018b,Mohammad2025} and NRC-EIL \cite{Mohammad2018a} to reduce ambiguity in syntax and better match the conditions under which the corresponding theories have been studied.
Besides, GoEmotions \cite{Demszky2020} provides a more conservative perspective, as it contains full sentences instead of single- and multi-word samples.

\begin{figure}[!tb]
    \centering

    \includegraphics[width=\linewidth]{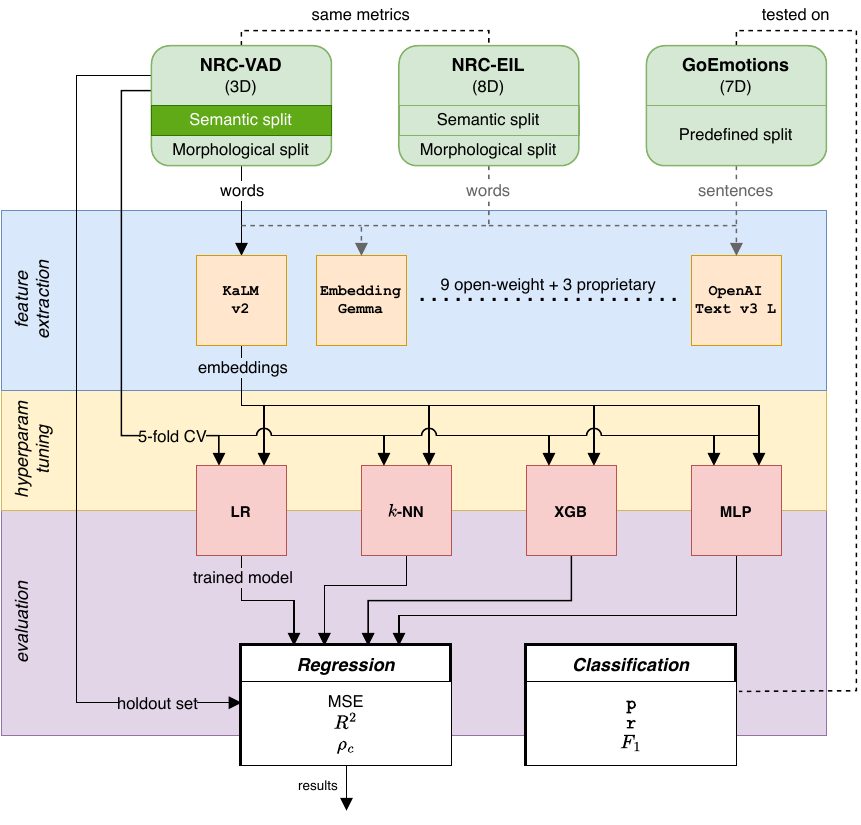}
    \caption{
        Pipeline demonstrating the fitting and evaluating procedure.
        All embeddings are calculated once and frozen for each dataset (blue section).
        For simplicity, the remaining control flow is depicted for one experiment only (yellow and purple sections), \ie the regression task of NRC-VAD in combination with semantic leakage prevention and using \texttt{KaLM v2} as text encoder.
    }
    \label{fig:system}
\end{figure}

We evaluated twelve text encoders (\cf \Cref{sec:models}) using a two-step procedure.
First, embeddings for all words and sentences in the datasets were computed and frozen.
Second, a collection of downstream predictors was trained and its hyperparameters were tuned, with the generated latent features as input, to assess the predictive performance on the corresponding regression and classification tasks (\cf \Cref{fig:system}).
We selected four predictors owning distinct characteristics to map embeddings and emotions, allowing to test whether emotional cues are accessible linearly or via nonlinear transformations (\cf \Cref{sec:predictors}).

To better measure the true generalization capabilities on the structured lexica, we applied two techniques to prevent morphological and semantic leakage across data splits, reducing evaluation biases on predictive models that rely on closely related lexical items to build their solutions.
The results presented in the main body of this paper focus on the semantics-aware splitting strategy only, as it is inherently less biased.
More comprehensive summaries, including those obtained with the morphology-aware approach, are reported in the Appendix (\cf \Cref{appx:split,appx:full}).

\subsection{Emotion datasets}\label{sec:data}

All corpora are freely accessible and in English.
They present varying input granularity (single-word, multi-word, and sentence-level) as well as output formats (discrete and continuous).

\begin{description}
    \item[NRC-VAD]
          \cite{Mohammad2018b,Mohammad2025} consists of around 55k single- and multi-word samples, annotated via crowdsourcing with real-valued valence, arousal, and dominance scores in the interval $[-1, 1]$, following Mehrabian's theory \cite{Mehrabian1974}.
    \item[NRC-EIL]
          \cite{Mohammad2018a} contains almost 6k single words with emotion intensities in the interval $[0, 1]$, in line with Plutchik's model \cite{Plutchik1980}.
          The properties of the collection would make it suitable as both a regression and a classification task, since 62.4\% of the entries are assigned a single emotion, 18.6\% have a pair of nonzero intensities, and the remaining terms are characterized by three or more emotions.
          In our experiments, we focused on regression of the real-valued intensities.
    \item[GoEmotions] \cite{Demszky2020} comprises over 54k comments crawled from Reddit, paired with 27 labels and filtered according to inter-rater agreement.
          The dataset provides official documentation to map these categories to a subset of labels consistent with Ekman's framework \cite{Ekman1971}.
          For our evaluations, we took the labeled sentences in combination with this projection to obtain a multi-label classification dataset with 7 classes, six for Ekman's emotions and one for the neutral category, where 91.2\% of the samples have one category and 8.8\% at least two.
\end{description}

\paragraph{Splitting strategy}
To train the predictive models, each dataset is partitioned into five folds for cross-validation and one holdout test set for final evaluation in a 80\%/20\% proportion.
Given that GoEmotions has a predefined train-dev-test split, we combined the training and development splits and applied stratified sampling to balance the labels in the folds, while the test split was left unchanged to enable comparisons with the evaluations of the original work.
For the two lexicon-based datasets, \ie NRC-VAD and NRC-EIL, we used a novel technique for semantic leakage prevention, as detailed in the next paragraph.

\paragraph{Leakage prevention}\label{par:leak}
Random train-test splits can overestimate generalization due to morphological or semantic leakage.
For instance, NRC-VAD includes multiple inflected and derived forms of the same lexical root (\eg \emph{pleasure}, \emph{pleasures}, \emph{pleasurable}) and semantically related terms (\eg \emph{calm}, \emph{chill}, \emph{peaceful}).
Text encoders tend to map these elements into nearby regions of the embedding space, which downstream predictive models could exploit by relying on nearest-neighbor similarity rather than learning to genuinely generalize.

To prevent this, we created a graph representation of the dataset lexemes and clustered them with the Leiden algorithm \cite{Traag2019}.
Nodes correspond to lexemes, with edges that are present if exceeding a threshold and weighted by the Wu--Palmer semantic similarity \cite{Wu1994} computed between the corresponding term synsets in WordNet \cite{Miller1995,Fellbaum1998}.
Lexemes not covered by WordNet are excluded from the dataset.

To assign the clusters to the cross-validation folds and the holdout set, we applied a greedy balancing algorithm that optimizes two criteria: (i) the split sizes, \ie 16\% for each fold and 20\% for the test set; and (ii) the preservation of the mean of the distribution for each emotion dimension across splits.
The procedure is executed multiple times with different initialization seeds and the best balanced split is kept.

\subsection{Text encoders}\label{sec:models}

For a comprehensive analysis, we handpicked six open-weight and three proprietary models from the top entries of the Massive Text Embedding Benchmark (MTEB)\footnote{Online at \url{https://huggingface.co/spaces/mteb/leaderboard}.} leaderboard curated by Hugging Face \cite{Muennighoff2023,Enevoldsen2025}.
When multiple models shared the same base architecture, including updated versions, we selected the encoder with the best average score.
Of the six considered open models, five of them (\texttt{KaLM Embedding v2}, \texttt{Qwen3 Embedding 8B}, \texttt{Linq Embed Mistral}, \texttt{LLaMA Embed Nemotron 8B}, and \texttt{Multilingual E5 Large Instruct}) are instruction-aware, \ie they were trained to generate embeddings optimized for user-defined specifications included as additional input \cite{Su2023}.
Differently, \texttt{EmbeddingGemma} is the only one which is task-tuned, \ie a relaxed instruction-aware model with predefined configurations for a set of use cases (\eg classification or semantic text similarity).
These were set up following the recommended prompts in their model cards together with the task to be solved, \ie regression and classification, to detail the desired feature extraction (\cf \Cref{appx:prompts}).
As for proprietary models, we chose \texttt{OpenAI Text Embedding v3 Large}, \texttt{Gemini Embedding 001}, and \texttt{Voyage v3 Large}, none of which is instruction-aware.

To further diversify the model pool, we also included two very recent encoders, \ie \texttt{Jina Embeddings v4} and \texttt{Nomic Embed v2}, whose earlier releases demonstrated good performance on the MTEB benchmark, but which have been assessed on a limited subset of the list of tasks in MTEB in their current versions.
Additionally, \texttt{Sentence T5 XXL}, a popular sentence embedding model, is included to serve as a representative for text encoders without explicit support for custom prompts or task instructions.

In total, twelve models are considered, spanning a wide range of parameter sizes, output dimensionalities, and training corpora, both English-only and multilingual (\cf \Cref{tab:models}).

\begingroup

\setcounter{footnote}{0}
\renewcommand*{\thefootnote}{\alph{footnote}}

\begin{table}[!tb]
    \centering

    \caption{
        Description of the analyzed text encoders.
        Licenses are subdivided into open-weight ({\small\faLockOpen}) and proprietary (\!{\small\faLock}\!).
        The number of parameters for downloadable models and the dimensionality of the latent features are specified by $p$ and $d$, respectively.
        Each entry is tagged with its type between no prompt support (\ding{115}), task-tuned ($\mspace{1mu}$\ding{110}$\mspace{1mu}$), and instruction-aware (\ding{108}).
        All encoders are multilingual, unless their training corpora are mainly in English ($\dagger$).
    }
    \label{tab:models}
    \begin{tabularx}{\linewidth}{l>{\centering\arraybackslash}X>{\centering\arraybackslash}X>{\centering\arraybackslash}X>{\centering\arraybackslash}X}
        \hline
        \textbf{Name}                                                                                                                              & \textbf{$p$} & \textbf{$d$} & \textbf{Access}     & \textbf{Reference}        \\
        \hline
        \href{https://huggingface.co/sentence-transformers/sentence-t5-xxl}{\texttt{Sentence T5 XXL}}\textsuperscript{\ding{115}$\dagger$}         & 4.8B         & $768$        & {\small\faLockOpen} & \cite{Ni2021}             \\
        \href{https://huggingface.co/google/embeddinggemma-300m}{\texttt{EmbeddingGemma}}\textsuperscript{\ding{110}}                              & 300M         & $768$        & {\small\faLockOpen} & \cite{Vera_Schechter2025} \\
        \href{https://huggingface.co/nomic-ai/nomic-embed-text-v2-moe}{\texttt{Nomic Embed v2}}\textsuperscript{\ding{110}}                        & 305M         & $768$        & {\small\faLockOpen} & \cite{Nussbaum2025}       \\
        \href{https://huggingface.co/intfloat/multilingual-e5-large-instruct}{\texttt{Multilingual E5 Large Instruct}}\textsuperscript{\ding{108}} & 560M         & $1024$       & {\small\faLockOpen} & \cite{Wang2024}           \\
        \href{https://huggingface.co/jinaai/jina-embeddings-v4}{\texttt{Jina Embeddings v4}}\textsuperscript{\ding{110}}                           & 3.8B         & $2048$       & {\small\faLockOpen} & \cite{Gunther2025}        \\
        \href{https://huggingface.co/tencent/KaLM-Embedding-Gemma3-12B-2511}{\texttt{KaLM Embedding v2}}\textsuperscript{\ding{108}}               & 12B          & $3840$       & {\small\faLockOpen} & \cite{Zhao2025}           \\
        \href{https://huggingface.co/Linq-AI-Research/Linq-Embed-Mistral}{\texttt{Linq Embed Mistral}}\textsuperscript{\ding{108}$\dagger$}        & 7B           & $4096$       & {\small\faLockOpen} & \cite{Junseong2024}       \\
        \href{https://huggingface.co/nvidia/llama-embed-nemotron-8b}{\texttt{LLaMA Embed Nemotron 8B}}\textsuperscript{\ding{108}}                 & 8B           & $4096$       & {\small\faLockOpen} & \cite{Babakhin2025}       \\
        \href{https://huggingface.co/Qwen/Qwen3-Embedding-8B}{\texttt{Qwen3 Embedding 8B}}\textsuperscript{\ding{108}}                             & 8B           & $4096$       & {\small\faLockOpen} & \cite{Zhang2025}          \\
        \texttt{Voyage v3 Large}\textsuperscript{\ding{115},}\hyperlink{note:voyage}{\footnotemark}                                                                         &              & $2048$       & {\small\faLock}                                 \\ 
        \texttt{OpenAI Text Embedding v3 Large}\textsuperscript{\ding{115},}\hyperlink{note:openai}{\footnotemark}                                                          &              & $3072$       & {\small\faLock}                                 \\
        \texttt{Gemini Embedding 001}\textsuperscript{\ding{110}}                                                                                  &              & $3072$       & {\small\faLock}     & \cite{Lee2025b}           \\
        \hline
    \end{tabularx}
\end{table}

\addtocounter{footnote}{-1}
\hypertarget{note:voyage}{\footnotetext{\url{https://blog.voyageai.com/2025/01/07/voyage-3-large}.}}
\addtocounter{footnote}{1}
\hypertarget{note:openai}{\footnotetext{\url{https://platform.openai.com/docs/models/text-embedding-3-large}.}}

\endgroup

\subsection{Predictive models}\label{sec:predictors}
We employed four predictors, \ie linear and logistic regression with elastic net regularization (LR), $k$-nearest neighbors ($k$-NN), XGBoost (XGB), and multilayer perceptron (MLP), to leverage text embeddings for downstream emotion prediction tasks.

On the regression datasets, \ie NRC-VAD and NRC-EIL, all predictive models were trained to minimize the mean squared error (MSE).
For multi-label classification, \ie GoEmotions, performance was optimized maximizing the macro-averaged $F_{1}$, using a fixed decision threshold of $0.5$.
Since LR and XGB do not natively support multi-label classification, one-vs-the-rest strategy was applied.
The hyperparameters of the predictors were independently tuned for each dataset and generating encoder over multiple optimization trials with respect to $R^2$ for regression and macro $F_{1}$ for classification.
More details on how hyperparameter tuning was carried out can be found in \Cref{appx:hyper}.

\section{Experimental results}\label{sec:results}

\subsection{Quantitative regression analysis}\label{sec:regr}

We evaluated the performance on the holdout test sets using three regression metrics, \ie MSE, $R^{2}$, and concordance correlation coefficient
\begin{equation}\label{eq.ccc}
    \rho_{c} = \frac{2 \rho\sigma_{x}\sigma_{y}}{\sigma_{x}^{2} + \sigma_{y}^{2} + (\mu_{x} - \mu_{y})^{2}}.
\end{equation}
In Equation~\ref{eq.ccc}, $x$ and $y$ denote true and predicted values respectively, whereas $\rho$ equals to Pearson's correlation coefficient.
$\rho_{c}$ was chosen because, as defined in its formula, it captures both correlation and agreement, penalizing scale mismatches \cite{Lin1989}.
To further support our results, the outcomes of paired difference tests, estimated via bootstrapping, are provided for each metric and encoder.
The null hypothesis claims that, using a given predictor as backend, the candidate encoder leads to better performance than the best encoder.

\Cref{tab:NRC-VAD,tab:NRC-EIL} sum up the inference performance, with the best results in bold and highlighting the cases without statistical evidence ($p > 0.05$) with a double underline and statistically significant $p$-values falling within the interval $[0.005, 0.05]$ with a single underline.
The absence of an underlined value refers to $p$-values below $0.005$.

\begin{table}[!tb]
    \centering

    \caption{Regression metrics at test time for NRC-VAD, sorted by $R^{2}$ score of the MLP model in descending order.}
    \label{tab:NRC-VAD}
    \begin{tabular}{l*{12}{c}}
                                             & \multicolumn{3}{c}{\textbf{LR}} & \multicolumn{3}{c}{$\bm{k}$\textbf{-NN}} & \multicolumn{3}{c}{\textbf{XGB}} & \multicolumn{3}{c}{\textbf{MLP}}                                                                                                                                                                      \\
                                             & MSE                             & $R^2$                                    & $\rho_{c}$                       & MSE                              & $R^2$              & $\rho_{c}$                     & MSE             & $R^2$           & $\rho_{c}$         & MSE             & $R^2$           & $\rho_{c}$      \\
        \cmidrule(l{1.75pt}r{1.75pt}){2-4}\cmidrule(l{1.75pt}r{1.75pt}){5-7}\cmidrule(l{1.75pt}r{1.75pt}){8-10}\cmidrule(l{1.75pt}r{1.75pt}){11-13}
        \textbf{\texttt{KaLM v2}}            & $\mathbf{.066}$                 & $\mathbf{.637}$                          & $\mathbf{.790}$                  & $\mathbf{.075}$                  & $\mathbf{.591}$    & $\underline{\underline{.746}}$ & $\mathbf{.068}$ & $\mathbf{.630}$ & $\mathbf{.777}$    & $\mathbf{.059}$ & $\mathbf{.677}$ & $\mathbf{.811}$ \\
        \textbf{\texttt{Linq Mistral}}       & $.067$                          & $.631$                                   & $.786$                           & $\underline{.076}$               & $\underline{.586}$ & $\mathbf{.746}$                & $.069$          & $.622$          & $\underline{.773}$ & $.061$          & $.667$          & $.806$          \\
        \textbf{\texttt{OpenAI Text v3 L}}   & $.073$                          & $.602$                                   & $.761$                           & $.079$                           & $.572$             & $.704$                         & $.077$          & $.582$          & $.734$             & $.062$          & $.659$          & $.797$          \\
        \textbf{\texttt{Qwen3 8B}}           & $.069$                          & $.624$                                   & $.780$                           & $.077$                           & $.578$             & $.733$                         & $.072$          & $.607$          & $.761$             & $.063$          & $.657$          & $.798$          \\
        \textbf{\texttt{LLaMA Nemotron 8B}}  & $.075$                          & $.589$                                   & $.745$                           & $.099$                           & $.463$             & $.607$                         & $.079$          & $.570$          & $.729$             & $.064$          & $.653$          & $.795$          \\
        \textbf{\texttt{Gemini 001}}         & $.077$                          & $.578$                                   & $.748$                           & $.080$                           & $.564$             & $.710$                         & $.081$          & $.560$          & $.717$             & $.066$          & $.638$          & $.782$          \\
        \textbf{\texttt{ST5 XXL}}            & $.080$                          & $.564$                                   & $.735$                           & $.077$                           & $.581$             & $.722$                         & $.080$          & $.564$          & $.715$             & $.068$          & $.631$          & $.777$          \\
        \textbf{\texttt{EmbeddingGemma}}     & $.082$                          & $.553$                                   & $.725$                           & $.085$                           & $.536$             & $.697$                         & $.082$          & $.553$          & $.713$             & $.073$          & $.603$          & $.757$          \\
        \textbf{\texttt{Voyage v3 L}}        & $.082$                          & $.556$                                   & $.728$                           & $.098$                           & $.472$             & $.624$                         & $.086$          & $.534$          & $.699$             & $.073$          & $.601$          & $.755$          \\
        \textbf{\texttt{Multilang E5 L Ins}} & $.084$                          & $.542$                                   & $.717$                           & $.085$                           & $.537$             & $.704$                         & $.083$          & $.550$          & $.713$             & $.077$          & $.582$          & $.741$          \\
        \textbf{\texttt{Jina v4}}            & $.088$                          & $.520$                                   & $.694$                           & $.096$                           & $.479$             & $.636$                         & $.091$          & $.506$          & $.671$             & $.080$          & $.563$          & $.726$          \\
        \textbf{\texttt{Nomic v2}}           & $.108$                          & $.413$                                   & $.603$                           & $.112$                           & $.396$             & $.540$                         & $.110$          & $.403$          & $.567$             & $.097$          & $.476$          & $.647$          \\
    \end{tabular}
\end{table}

\subsubsection{NRC-VAD}

\texttt{KaLM v2} achieves the highest scores across all predictive models with one exception, \ie $\rho_c$ for $k$-NN, where \texttt{Linq Mistral} is slightly more performant, though not in a statistically significant way (\cf double-underlined value in the first row of \Cref{tab:NRC-VAD}).
\texttt{OpenAI Text v3 L} ranks third with the MLP backend, whereas other instruction-aware models, \ie \texttt{Qwen3 8B} and \texttt{LLaMA Nemotron 8B}, show comparable performance, followed by the task-tuned \texttt{Gemini 001}.
In relation to \descref{3}, these insights indicate that some open-weight text encoders can significantly outperform proprietary alternatives (\cf no underlined scores for \texttt{OpenAI Text v3 L}), likely due to instruction-awareness.
This also provides initial evidence for \descref{2}. Interestingly, the remaining instructional \texttt{Multilang E5 L Ins} occupies one of the last positions.

The maximum $R^2$ score of $.677$ and correlation $\rho$ of $.811$ indicate that the text encoders are capable of capturing affective cues, thereby addressing \descref{1}.
In general, with the MLP predictor, all encoders except for \texttt{Nomic v2} reach a $R^2 > .55$ and $\rho > .7$, hinting their ability to enclose affective signals.
Concerning \descref{4}, the extraction of this affective information is more effective when using the MLP backend, with LR as runner-up.

\begin{table}[!tb]
    \centering

    \caption{Regression metrics at test time for NRC-EIL, sorted by $R^{2}$ score of the MLP model in descending order.}
    \label{tab:NRC-EIL}
    \begin{tabular}{l*{12}{c}}
                                             & \multicolumn{3}{c}{\textbf{LR}} & \multicolumn{3}{c}{$\bm{k}$\textbf{-NN}} & \multicolumn{3}{c}{\textbf{XGB}} & \multicolumn{3}{c}{\textbf{MLP}}                                                                                                                                                                                                                                                 \\
                                             & MSE                             & $R^2$                                    & $\rho_{c}$                       & MSE                              & $R^2$                          & $\rho_{c}$         & MSE                            & $R^2$                          & $\rho_{c}$                     & MSE                            & $R^2$              & $\rho_{c}$                     \\
        \cmidrule(l{1.75pt}r{1.75pt}){2-4}\cmidrule(l{1.75pt}r{1.75pt}){5-7}\cmidrule(l{1.75pt}r{1.75pt}){8-10}\cmidrule(l{1.75pt}r{1.75pt}){11-13}
        \textbf{\texttt{KaLM v2}}            & $\underline{\underline{.023}}$  & $\underline{\underline{.516}}$           & $\underline{\underline{.705}}$   & $\mathbf{.024}$                  & $\underline{\underline{.500}}$ & $\underline{.685}$ & $\underline{\underline{.023}}$ & $\underline{\underline{.515}}$ & $\underline{\underline{.697}}$ & $\mathbf{.022}$                & $\mathbf{.540}$    & $\mathbf{.730}$                \\
        \textbf{\texttt{Linq Mistral}}       & $\mathbf{.023}$                 & $\mathbf{.519}$                          & $\mathbf{.710}$                  & $\underline{\underline{.024}}$   & $\mathbf{.500}$                & $\mathbf{.695}$    & $\mathbf{.023}$                & $\mathbf{.517}$                & $\mathbf{.704}$                & $\underline{\underline{.022}}$ & $\underline{.528}$ & $\underline{\underline{.729}}$ \\
        \textbf{\texttt{Qwen3 8B}}           & $\underline{\underline{.023}}$  & $\underline{\underline{.517}}$           & $\underline{\underline{.704}}$   & $\underline{\underline{.024}}$   & $\underline{\underline{.491}}$ & $\underline{.683}$ & $\underline{\underline{.023}}$ & $\underline{\underline{.513}}$ & $\underline{\underline{.698}}$ & $\underline{.022}$             & $\underline{.524}$ & $\underline{\underline{.725}}$ \\
        \textbf{\texttt{EmbeddingGemma}}     & $.024$                          & $.491$                                   & $.683$                           & $\underline{.024}$               & $\underline{.486}$             & $.672$             & $.024$                         & $.485$                         & $.667$                         & $\underline{.022}$             & $\underline{.524}$ & $.717$                         \\
        \textbf{\texttt{LLaMA Nemotron 8B}}  & $.025$                          & $.475$                                   & $.673$                           & $.026$                           & $.446$                         & $.647$             & $.026$                         & $.456$                         & $.637$                         & $\underline{.023}$             & $\underline{.523}$ & $\underline{.720}$             \\
        \textbf{\texttt{OpenAI Text v3 L}}   & $.025$                          & $.471$                                   & $.658$                           & $.026$                           & $.454$                         & $.631$             & $.026$                         & $.445$                         & $.616$                         & $.023$                         & $.516$             & $.704$                         \\
        \textbf{\texttt{ST5 XXL}}            & $.025$                          & $.471$                                   & $.659$                           & $.025$                           & $.474$                         & $.659$             & $.026$                         & $.449$                         & $.626$                         & $.023$                         & $.513$             & $.712$                         \\
        \textbf{\texttt{Multilang E5 L Ins}} & $.025$                          & $.472$                                   & $.671$                           & $.025$                           & $.472$                         & $.665$             & $.024$                         & $.489$                         & $.676$                         & $.023$                         & $.508$             & $.709$                         \\
        \textbf{\texttt{Gemini 001}}         & $.026$                          & $.460$                                   & $.654$                           & $.026$                           & $.458$                         & $.648$             & $.027$                         & $.436$                         & $.613$                         & $.024$                         & $.499$             & $.699$                         \\
        \textbf{\texttt{Voyage v3 L}}        & $.026$                          & $.446$                                   & $.645$                           & $.029$                           & $.401$                         & $.578$             & $.027$                         & $.427$                         & $.605$                         & $.024$                         & $.488$             & $.688$                         \\
        \textbf{\texttt{Jina v4}}            & $.028$                          & $.408$                                   & $.600$                           & $.030$                           & $.372$                         & $.558$             & $.029$                         & $.384$                         & $.572$                         & $.027$                         & $.432$             & $.631$                         \\
        \textbf{\texttt{Nomic v2}}           & $.032$                          & $.336$                                   & $.529$                           & $.034$                           & $.293$                         & $.457$             & $.033$                         & $.313$                         & $.479$                         & $.030$                         & $.369$             & $.578$                         \\
    \end{tabular}
\end{table}

\subsubsection{NRC-EIL}

Similar patterns emerge with respect to NRC-VAD.
\texttt{KaLM v2} and \texttt{Linq Mistral} consistently occupy the top positions.
\texttt{KaLM v2} benefits again from the MLP predictor (cf. top-right corner of \Cref{tab:NRC-EIL}), while \texttt{Linq Mistral} performs better across almost all other predictive models (\cf bold scores on second row).
In contrast to the ranking on NRC-VAD, \texttt{Qwen3 8B} and \texttt{EmbeddingGemma} rise to the third and the fourth places, while \texttt{OpenAI Text v3 L} drops to the sixth position.
This addresses \descref{4}.

Performance differences between \texttt{Linq Mistral} and \texttt{KaLM v2}, \texttt{Qwen3 8B}, and \texttt{EmbeddingGemma} have $p$-values in $[0.005, 0.05]$, indicating that the ranking is not statistically decisive (\cf single-underlined values in the corresponding rows).
This suggests that task-tuned encoders such as \texttt{EmbeddingGemma} are competitive with respect to instruction-aware alternatives, providing further insight into \descref{2}.
With regard to \descref{3}, four instruction-aware models and one task-tuned encoder outperform all proprietary models (\cf top-5 ranking and rows below), hence the support for generating optimized embeddings via instructions appears to be generally beneficial.

As for \descref{1}, the highest $R^2$ reaches a score of $.540$ and $\rho$ a correlation of $.730$, showing a moderate encoding of affective information for the top performing embedding model.
Recent open encoders, \ie \texttt{Jina v4} and \texttt{Nomic v2}, seem to weakly enclose affective cues, as hinted by their low $R^2$ scores of $.432$ and $.369$ (\cf last rows).

\begin{table}[!tb]
    \centering

    \caption{Summary of the macro-averaged classification metrics at test time for GoEmotions, sorted by $F_{1}$ score of the MLP model in descending order.}
    \label{tab:GoEmotions}
    \begin{tabular}{l*{12}{c}}
                                             & \multicolumn{3}{c}{\textbf{LR}} & \multicolumn{3}{c}{$\bm{k}$\textbf{-NN}} & \multicolumn{3}{c}{\textbf{XGB}} & \multicolumn{3}{c}{\textbf{MLP}}                                                                                                                                                                                                                                           \\
                                             & \texttt{p}                      & \texttt{r}                               & $F_{1}$                          & \texttt{p}                       & \texttt{r}      & $F_{1}$         & \texttt{p}                     & \texttt{r}                     & $F_{1}$                        & \texttt{p}                     & \texttt{r}                     & $F_{1}$                        \\
        \cmidrule(l{1.75pt}r{1.75pt}){2-4}\cmidrule(l{1.75pt}r{1.75pt}){5-7}\cmidrule(l{1.75pt}r{1.75pt}){8-10}\cmidrule(l{1.75pt}r{1.75pt}){11-13}
        \textbf{\texttt{Gemini 001}}         & $\mathbf{.716}$                 & $\mathbf{.517}$                          & $\mathbf{.594}$                  & $\mathbf{.624}$                  & $\mathbf{.543}$ & $\mathbf{.575}$ & $\underline{\underline{.674}}$ & $\mathbf{.469}$                & $\mathbf{.546}$                & $\mathbf{.713}$                & $\mathbf{.529}$                & $\mathbf{.600}$                \\
        \textbf{\texttt{EmbeddingGemma}}     & $\underline{.688}$              & $.493$                                   & $.568$                           & $\underline{.599}$               & $.512$          & $.546$          & $\underline{\underline{.678}}$ & $\underline{.447}$             & $\underline{\underline{.533}}$ & $\underline{\underline{.700}}$ & $\underline{\underline{.517}}$ & $\underline{\underline{.590}}$ \\
        \textbf{\texttt{OpenAI Text v3 L}}   & $\underline{.687}$              & $.489$                                   & $.564$                           & $.472$                           & $.423$          & $.437$          & $\underline{.645}$             & $.397$                         & $.483$                         & $.683$                         & $\underline{\underline{.518}}$ & $\underline{.579}$             \\
        \textbf{\texttt{Linq Mistral}}       & $\underline{.685}$              & $\underline{.494}$                       & $.568$                           & $.565$                           & $.461$          & $.500$          & $\underline{\underline{.666}}$ & $.408$                         & $.497$                         & $\underline{\underline{.700}}$ & $.494$                         & $.575$                         \\
        \textbf{\texttt{Qwen3 8B}}           & $\underline{\underline{.700}}$  & $\underline{.496}$                       & $\underline{.573}$               & $.582$                           & $.490$          & $.525$          & $\underline{.651}$             & $.425$                         & $.507$                         & $\underline{\underline{.703}}$ & $.498$                         & $.574$                         \\
        \textbf{\texttt{KaLM v2}}            & $\underline{\underline{.705}}$  & $\underline{\underline{.515}}$           & $\underline{\underline{.588}}$   & $\underline{.600}$               & $.511$          & $.544$          & $\underline{\underline{.682}}$ & $\underline{\underline{.453}}$ & $\underline{\underline{.535}}$ & $\underline{\underline{.704}}$ & $.487$                         & $.565$                         \\
        \textbf{\texttt{Multilang E5 L Ins}} & $.674$                          & $.474$                                   & $.549$                           & $.511$                           & $.496$          & $.499$          & $\underline{\underline{.674}}$ & $.392$                         & $.481$                         & $.675$                         & $.482$                         & $.550$                         \\
        \textbf{\texttt{LLaMA Nemotron 8B}}  & $.637$                          & $.454$                                   & $.521$                           & $.476$                           & $.405$          & $.425$          & $\mathbf{.683}$                & $.388$                         & $.480$                         & $.607$                         & $\underline{\underline{.512}}$ & $.549$                         \\
        \textbf{\texttt{Nomic v2}}           & $.669$                          & $.428$                                   & $.514$                           & $.476$                           & $.410$          & $.429$          & $\underline{.639}$             & $.348$                         & $.438$                         & $.669$                         & $.460$                         & $.537$                         \\
        \textbf{\texttt{Jina v4}}            & $.663$                          & $.423$                                   & $.508$                           & $.454$                           & $.395$          & $.412$          & $.631$                         & $.346$                         & $.434$                         & $.660$                         & $.436$                         & $.515$                         \\
        \textbf{\texttt{Voyage v3 L}}        & $.635$                          & $.420$                                   & $.498$                           & $.442$                           & $.382$          & $.399$          & $.568$                         & $.327$                         & $.406$                         & $.621$                         & $.447$                         & $.513$                         \\
        \textbf{\texttt{ST5 XXL}}            & $.649$                          & $.424$                                   & $.506$                           & $.449$                           & $.408$          & $.420$          & $.618$                         & $.350$                         & $.436$                         & $.668$                         & $.427$                         & $.510$                         \\
    \end{tabular}
\end{table}

\subsection{Quantitative classification analysis}

We used three common measures, \ie precision, recall, and $F_{1}$ score, aggregating with respect to multiple classes by calculating the macro average of these metrics.
\Cref{tab:GoEmotions} summarizes the aggregated results.
As in \Cref{sec:regr}, results are supported by significance checks.
In particular, paired permutation tests were applied.

\subsubsection{GoEmotions}

Results reveal a substantially different ranking in comparison with the regression analysis.
The proprietary \texttt{Gemini 001} and the open-weight \texttt{EmbeddingGemma} achieve the top places on all predictive backends with their $F_1$ scores, followed by \texttt{OpenAI Text v3 L} (\cf top-3 ranking of \Cref{tab:GoEmotions}).
Interestingly, \texttt{KaLM v2} achieves an $F_1$ score of $.588$ with LR, which is the only occurrence where MLP is outperformed (\cf $F_1$ score of $.565$ in the sixth row).
This makes \texttt{\texttt{KaLM v2}} the model with the third best $F_1$, giving new insights on \descref{4}.
Two open encoders without explicit prompt support for optimized embeddings, \ie \texttt{Voyage v3 L} and \texttt{ST5 XXL}, occupy the last positions (\cf last rows).

Considering \descref{3}, in contrast to the analyses on NRC-VAD and NRC-EIL, the top ranker is a proprietary text encoder, \ie \texttt{Gemini 001}, rather than an open-weight model.
However, this superior performance is not statistically significant (\cf underlined $F_1$ scores for \texttt{EmbeddingGemma}).
In relation to \descref{2}, this highlights that instruction-aware encoders, as well as models with large parameter size and embedding dimensionality, are not always better than task-tuned or proprietary alternatives, in particular when sentence-level samples are used instead of word-level data.

Concerning \descref{1}, a maximum $F_1$ of $0.60$ is achieved (\cf MLP column in the top-right corner).
Instead, the authors of the dataset report a score of $0.64$ by fine-tuning \texttt{BERT} \cite{Demszky2020}, though without specifying the selected classification threshold.
This comparison suggests that exploiting fine-tuning could be advantageous.

As for \descref{4}, in terms of $F_1$ scores, the MLP backend consistently outperforms all other predictive models, with only one exception (\cf LR backend for \texttt{KaLM v2}).
LR ranks second, whereas $k$-NN and XGB exhibit more variable performance depending on the used encoder.
These trends are in line with those observed in the experiments on regression.

\subsection{Qualitative visual analysis}

\begin{figure}[!tb]
    \centering

    \includegraphics[width=\linewidth]{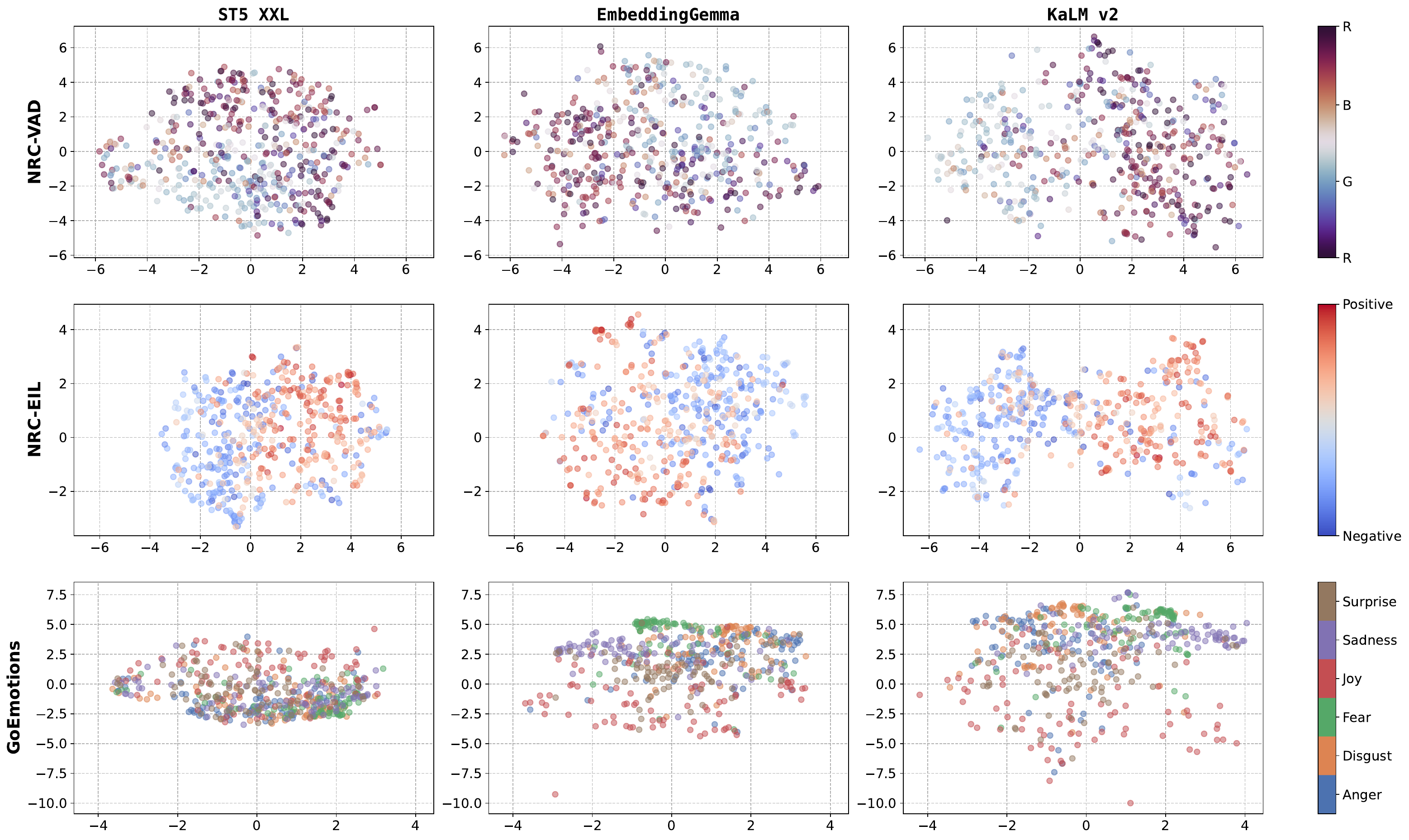}
    \caption{UMAP visualization of the full embeddings with color-coded labels.}
    \label{fig:viz}
\end{figure}

We report a series of visualizations with the intent to discover whether generated vectors are expressive enough to imply a clustering of similar elements with respect to their affect.
For this analysis, we focus on the best performing open-weight representatives for each of the three types of text encoder, \ie without prompt support, task-tuned, and instruction-aware, as specified in \Cref{tab:models}.

The embeddings calculated on the full datasets were transformed into a 2D representation with UMAP \cite{McInnes2018} set up with cosine similarity as metric to compare vectors.
Depending on the data format, the output variable information was converted through a color encoding as follows.
\begin{itemize}
    \item 3D emotion points from NRC-VAD were read as RGB triples and their hue value was used as parametrization of a cyclic colormap. Samples with pure valence (R), arousal (G), or dominance (B) signals are equidistant, with respect to the color space, to other pure points.
    \item Entries of NRC-EIL with more than one active affect intensity were filtered out and the remaining elements were transmuted into members of the positive emotions (\textit{joy}, \textit{trust}, \textit{anger}, and \textit{anticipation}) or negative counterparts (\textit{sadness}, \textit{disgust}, \textit{fear}, and \textit{surprise}) following Plutchik's original statement, where the two groups render the opposite ends of a diverging colormap.
    \item Samples from GoEmotions either having multiple labels or tagged as neutral were dropped and linked to Ekman's taxonomy through the official dataset lookup table, with each category corresponding to a distinct color.
\end{itemize}

For the sake of readability, \Cref{fig:viz} is limited to 500 points per dataset, where stratified sampling was applied to NRC-EIL and GoEmotions.
In the first row, it can be seen that none of the encoders is able to entail the creation of clusters of consistent entries for NRC-VAD.
In fact, an equilateral triangle of uniformly distributed samples, with pure points of a component at its vertices, should ideally form.
Among the three affective axes, the one for arousal looks to be the most easily separable when considered alone.
Instead, all text encoders can sharply divide the elements of NRC-EIL into two groupings, as evident in the second line of plots.
In addition, the instruction-aware model is particularly skilled at avoiding poisoning the cluster of positive samples with intense negative points.
As for GoEmotions in the third row, while disgust-, fear-, and sadness-labeled entries tend to gather together, especially in the task-tuned and instruction-aware encoders, elements tagged with the anger, joy, and surprise labels spread out and hardly group.
The difficulty might be possibly due to the fact that semantically broader labels are intuitively associated with the tendency to be more frequently selected by a human rater.
Since a bigger cardinality for the subset of a category can imply a higher variance in its embedding vectors, samples marked with the most occurring tags have a higher risk of being characterized by inconsistent latent features with respect to a representative of their cluster of belongingness.

\section{Conclusions and future works}\label{sec:outro}

To conclude, our analyses show that affective information is present to a varying degree within twelve text encoders and across three emotion frameworks.
Addressing \descref{1}, the highest $R^2$ score achieved by the best encoder on VAD regression has been of $.677$, highlighting that affective information is well-represented by the embeddings within this framework.
In contrast, for Plutchik regression, the maximum score has been of $.540$, likely due to the higher dimensionality (\ie eight versus three) and the smaller dataset size.
As for the sentence-based multi-label classification over Ekman's six emotions plus a neutral one, the highest $F_1$ score has been of $.600$.
In addition, visualizing the down projections of the embeddings reveals similar patterns.
Concerning \descref{4}, affective cues appear to be more readily accessible through nonlinear downstream predictors, even though linear transformations can achieve comparable results.
On the lexicon datasets, instruction-aware models rank the highest and outperform other candidates, including proprietary ones.
However, this trend does not hold for sentence-level data, where task-tuned models are the best performing, giving insights into \descref{2} and \descref{3}.

Regarding future works, we pivoted on three emotion theories, considering both the categorical and the dimensional families.
However, other frameworks motivated by psychology and specifically developed for affective computing have been presented \cite{Cambria2012}.
Therefore, it would be desirable to extend our overview to them.
Additionally, we focused on discovering emotional cues from text.
Nevertheless, most of the latest LMs have been trained with multimodality in mind.
It has already been observed that incorporating different types of data can improve the understanding capabilities on a vast series of downstream tasks.\footnote{For instance, see the technical report at \url{https://www.anthropic.com/claude-3-model-card}.}
Consequently, adapting our methodology and applying it to multimodal resources \cite{Busso2008} could highlight even more which embedding models are the best in distinguishing the nuanced facets of emotions.

\section*{Limitations}

We should acknowledge that the prompts of the task-based and instruction-aware text encoders under examination were set up at our own discretion.
To enable fair evaluation and comparison, we tried to configure these models as much as possible with matching settings for feature extraction.
However, it is reasonable to suppose that, for each specific encoder and task between regression and classification, there might exist other instructions which would imply better predictive performance.



\bibliographystyle{splncs04}
\bibliography{references}

\renewcommand{\theHsection}{A\arabic{section}}

\appendix

\begingroup
    \hypersetup{bookmarksdepth=0}
    \section{Morphological split}\label{appx:split}

For NRC-VAD and NRC-EIL, along with the semantics-aware splitting strategy, we also applied a simpler approach to prevent morphological leakage between the 5-folds for cross-validation and the holdout test set.
By comparing the results of both strategies, the impact of allowing semantic information exchange across splits can be shown.

The morphological split was created by using stop word removal and Snowball stemming, assigning words with the same stem to the same group.
All words in a group need to be in the same split.
This is accomplished by the greedy algorithm in \Cref{par:leak}.

\section{Prompts}\label{appx:prompts}

In \Cref{listing:prompts}, we detail a subset of the prompts used to instantiate the analyzed task-tuned and instruction-aware text encoders.
As already mentioned in \Cref{sec:models}, these prompts specify the downstream problem under examination, \ie regression or classification, taking inspiration from the recommended prompts of each model card.

To be precise, for regression, we slightly modified default prompts tailored to either semantic text similarity (STS) or retrieval if no STS option was available.

\begin{listing*}[!tb]
    \caption{
        Configured prompts for \texttt{EmbeddingGemma}, \texttt{KaLM v2}, and \texttt{Linq Mistral}.
        The other text encoders are set up in an analogous way, with similar prompts with respect to this sample list.
        \texttt{prompt\_name} and \texttt{prompt} are arguments passed to the model instantiation via the Hugging Face API in Python.
    }
    \label{listing:prompts}
\begin{minted}{js}
{
    "google@embeddinggemma-300m": {
        "regression_configs": {
            "prompt_name": "STS"
        },
        "classification_configs": {
            "prompt_name": "Classification"
        }
    },
    "tencent@KaLM-Embedding-Gemma3-12B-2511": {
        "regression_configs": {
            "prompt": "Instruct: Retrieving emotion.\nQuery:"
        },
        "classification_configs": {
            "prompt": "Instruct: Classifying emotion.\nQuery:"
        }
    },
    "Linq-AI-Research@Linq-Embed-Mistral": {
        "regression_configs": {
            "prompt": "Instruct: Retrieve the emotion expressed in the text\nQuery: "
        },
        "classification_configs": {
            "prompt": "Instruct: Classify the emotion expressed in the text\nQuery: "
        }
    }
}
\end{minted}
\end{listing*}

\section{Hyperparameter tuning}\label{appx:hyper}

Hyperparameters were tuned for each predictive model ($4$), dataset and splitting strategy ($5$), and encoder ($12$), leading to $240$ experiments in total.
We chose Optuna \cite{Akiba2019} as Bayesian optimization procedure.

The set of tuneable hyperparameters excluded those that we expected to uniformly influence the performance of a given predictor.
For instance, the batch size of the MLP was kept fixed, as it mainly refers to training efficiency and is unlikely to influence how well affective cues are extracted from text embeddings.
The number of runs was heuristically determined with respect to the complexity of the search space under examination (\eg number of hyperparameters, continuous or categorical variables), while ensuring that all predictive models had sufficient chances to find their optimal configuration.

The final hyperparameters are listed from \Cref{tab:opt_first} to \Cref{tab:opt_last}.
Most of them are self-explanatory, except for the \texttt{complexity} parameter of the MLP, which controls the network architecture in terms of number and size of its hidden layers.
Each complexity level is a categorical variable, with twelve possible values, that corresponds to a predefined configuration, from shallow architectures with a single hidden layer (\eg $64$ neurons for complexity \#0 and $1024$ for \#4) to deeper networks (\eg one layer for levels from \#0 to \#4, two from \#5 to \#8, and three from \#9 to \#11).

\section{Full reports}\label{appx:full}

As supplementary documentation, we attach more detailed quantitative results.
Considering NRC-VAD (\cf \Cref{tab:NRC-VAD_Full}) and NRC-EIL (\cf \Cref{tab:NRC-EIL_Full}), the outcomes with the morphology-aware splitting strategy are added. For GoEmotions (\cf \Cref{tab:GoEmotions_Full}), we include weighted- and micro-averaged metrics as auxiliary measures for a more complete view.

Furthermore, we report mean and standard deviation of the cross-validation scores of the training procedure to give insights into the variability across folds (\cf \Cref{tab:NRC-VAD_Full_Metrics_1,tab:NRC-VAD_Full_Metrics_2,tab:NRC-EIL_Full_Metrics_1,tab:NRC-EIL_Full_Metrics_2,tab:GoEmotions_Full_Metrics_1,tab:GoEmotions_Full_Metrics_2}).

\begin{table}[!htb]
    \centering

    \caption{Summary of the best hyperparameters for LR on NRC-VAD, split with the semantics-aware strategy.}
    \label{tab:opt_first}
\begin{tabularx}{\linewidth}{lXXX}
        \toprule
                                             & $\bm{R^2}$ & \textbf{\texttt{alpha}} & \textbf{\texttt{l1\_ratio}} \\
        \midrule
        \textbf{\texttt{ST5 XXL}}            & 0.562      & 9.7e-04                 & 7.5e-04                     \\
        \textbf{\texttt{EmbeddingGemma}}     & 0.546      & 1.6e-03                 & 0.231                       \\
        \textbf{\texttt{Nomic v2}}           & 0.407      & 2.7e-03                 & 0.162                       \\
        \textbf{\texttt{Multilang E5 L Ins}} & 0.526      & 4.0e-04                 & 0.286                       \\
        \textbf{\texttt{Jina v4}}            & 0.516      & 1.7e-03                 & 0.475                       \\
        \textbf{\texttt{Linq Mistral}}       & 0.626      & 6.4e-03                 & 0.094                       \\
        \textbf{\texttt{KaLM v2}}            & 0.631      & 6.7e-04                 & 0.840                       \\
        \textbf{\texttt{LLaMA Nemotron 8B}}  & 0.582      & 0.030                   & 0.032                       \\
        \textbf{\texttt{Qwen3 8B}}           & 0.617      & 3.9e-03                 & 0.153                       \\
        \textbf{\texttt{Voyage v3 L}}        & 0.548      & 6.8e-04                 & 0.354                       \\
        \textbf{\texttt{OpenAI Text v3 L}}   & 0.597      & 1.2e-03                 & 0.729                       \\
        \textbf{\texttt{Gemini 001}}         & 0.575      & 2.6e-04                 & 0.403                       \\
        \bottomrule
    \end{tabularx}
\end{table}

\begin{table}[!htb]
    \centering

    \caption{Summary of the best hyperparameters for LR on NRC-VAD, split with the morphology-aware strategy.}
    \begin{tabularx}{\linewidth}{lXXX}
        \toprule
                                             & $\bm{R^2}$ & \textbf{\texttt{alpha}} & \textbf{\texttt{l1\_ratio}} \\
        \midrule
        \textbf{\texttt{ST5 XXL}}            & 0.567      & 4.3e-04                 & 5.3e-03                     \\
        \textbf{\texttt{EmbeddingGemma}}     & 0.544      & 1.0e-04                 & 0.026                       \\
        \textbf{\texttt{Nomic v2}}           & 0.407      & 1.7e-03                 & 0.216                       \\
        \textbf{\texttt{Multilang E5 L Ins}} & 0.531      & 1.1e-04                 & 0.910                       \\
        \textbf{\texttt{Jina v4}}            & 0.518      & 1.6e-03                 & 0.480                       \\
        \textbf{\texttt{Linq Mistral}}       & 0.637      & 5.4e-04                 & 0.743                       \\
        \textbf{\texttt{KaLM v2}}            & 0.640      & 0.027                   & 9.4e-03                     \\
        \textbf{\texttt{LLaMA Nemotron 8B}}  & 0.593      & 7.9e-03                 & 0.113                       \\
        \textbf{\texttt{Qwen3 8B}}           & 0.623      & 0.010                   & 0.038                       \\
        \textbf{\texttt{Voyage v3 L}}        & 0.549      & 2.0e-04                 & 0.800                       \\
        \textbf{\texttt{OpenAI Text v3 L}}   & 0.602      & 7.5e-04                 & 0.934                       \\
        \textbf{\texttt{Gemini 001}}         & 0.581      & 1.5e-03                 & 0.025                       \\
        \bottomrule
    \end{tabularx}
\end{table}

\begin{table}[!htb]
    \centering

    \caption{Summary of the best hyperparameters for LR on NRC-EIL, split with the semantics-aware strategy.}
    \begin{tabularx}{\linewidth}{lXXX}
        \toprule
                                             & $\bm{R^2}$ & \textbf{\texttt{alpha}} & \textbf{\texttt{l1\_ratio}} \\
        \midrule
        \textbf{\texttt{ST5 XXL}}            & 0.455      & 0.027                   & 0.140                       \\
        \textbf{\texttt{EmbeddingGemma}}     & 0.484      & 6.0e-03                 & 0.563                       \\
        \textbf{\texttt{Nomic v2}}           & 0.331      & 8.3e-03                 & 0.605                       \\
        \textbf{\texttt{Multilang E5 L Ins}} & 0.469      & 0.340                   & 3.3e-03                     \\
        \textbf{\texttt{Jina v4}}            & 0.401      & 0.200                   & 0.019                       \\
        \textbf{\texttt{Linq Mistral}}       & 0.515      & 6.8e-03                 & 0.610                       \\
        \textbf{\texttt{KaLM v2}}            & 0.505      & 4.8e-03                 & 0.864                       \\
        \textbf{\texttt{LLaMA Nemotron 8B}}  & 0.465      & 6.3e-03                 & 0.768                       \\
        \textbf{\texttt{Qwen3 8B}}           & 0.500      & 5.8e-03                 & 0.730                       \\
        \textbf{\texttt{Voyage v3 L}}        & 0.433      & 8.2e-03                 & 0.300                       \\
        \textbf{\texttt{OpenAI Text v3 L}}   & 0.461      & 0.011                   & 0.423                       \\
        \textbf{\texttt{Gemini 001}}         & 0.455      & 6.1e-03                 & 0.838                       \\
        \bottomrule
    \end{tabularx}
\end{table}

\begin{table}[!htb]
    \centering

    \caption{Summary of the best hyperparameters for LR on NRC-EIL, split with the morphology-aware strategy.}
    \begin{tabularx}{\linewidth}{lXXX}
        \toprule
                                             & $\bm{R^2}$ & \textbf{\texttt{alpha}} & \textbf{\texttt{l1\_ratio}} \\
        \midrule
        \textbf{\texttt{ST5 XXL}}            & 0.465      & 7.1e-03                 & 0.496                       \\
        \textbf{\texttt{EmbeddingGemma}}     & 0.488      & 0.043                   & 0.060                       \\
        \textbf{\texttt{Nomic v2}}           & 0.339      & 0.011                   & 0.412                       \\
        \textbf{\texttt{Multilang E5 L Ins}} & 0.469      & 6.5e-03                 & 0.371                       \\
        \textbf{\texttt{Jina v4}}            & 0.404      & 7.1e-03                 & 0.529                       \\
        \textbf{\texttt{Linq Mistral}}       & 0.524      & 0.048                   & 0.072                       \\
        \textbf{\texttt{KaLM v2}}            & 0.513      & 3.8e-03                 & 0.947                       \\
        \textbf{\texttt{LLaMA Nemotron 8B}}  & 0.465      & 0.026                   & 0.171                       \\
        \textbf{\texttt{Qwen3 8B}}           & 0.509      & 0.011                   & 0.358                       \\
        \textbf{\texttt{Voyage v3 L}}        & 0.444      & 0.034                   & 0.057                       \\
        \textbf{\texttt{OpenAI Text v3 L}}   & 0.475      & 0.015                   & 0.293                       \\
        \textbf{\texttt{Gemini 001}}         & 0.464      & 6.6e-03                 & 0.689                       \\
        \bottomrule
    \end{tabularx}
\end{table}

\begin{table}[!htb]
    \centering

    \caption{Summary of the best hyperparameters for LR on GoEmotions.}
    \begin{tabularx}{\linewidth}{lXXX}
        \toprule
                                             & $\bm{F_1}$ & \textbf{\texttt{C}} & \textbf{\texttt{l1\_ratio}} \\
        \midrule
        \textbf{\texttt{ST5 XXL}}            & 0.476      & 2.210               & 0.405                       \\
        \textbf{\texttt{EmbeddingGemma}}     & 0.559      & 0.168               & 0.625                       \\
        \textbf{\texttt{Nomic v2}}           & 0.496      & 0.288               & 0.241                       \\
        \textbf{\texttt{Multilang E5 L Ins}} & 0.517      & 43.115              & 0.693                       \\
        \textbf{\texttt{Jina v4}}            & 0.489      & 0.182               & 0.840                       \\
        \textbf{\texttt{Linq Mistral}}       & 0.544      & 0.074               & 0.215                       \\
        \textbf{\texttt{KaLM v2}}            & 0.566      & 0.012               & 2.1e-03                     \\
        \textbf{\texttt{LLaMA Nemotron 8B}}  & 0.509      & 0.178               & 0.767                       \\
        \textbf{\texttt{Qwen3 8B}}           & 0.551      & 0.111               & 0.962                       \\
        \textbf{\texttt{Voyage v3 L}}        & 0.490      & 0.724               & 0.806                       \\
        \textbf{\texttt{OpenAI Text v3 L}}   & 0.548      & 0.059               & 0.520                       \\
        \textbf{\texttt{Gemini 001}}         & 0.573      & 9.0e-03             & 0.087                       \\
        \bottomrule
    \end{tabularx}
\end{table}

\begin{table}[!htb]
    \centering

    \caption{Summary of the best hyperparameters for $k$-NN on NRC-VAD, split with the semantics-aware strategy.}
    \begin{tabularx}{\linewidth}{lXXX}
        \toprule
                                             & $\bm{R^2}$ & \textbf{\texttt{n\_neighbors}} & \textbf{\texttt{weights}} \\
        \midrule
        \textbf{\texttt{ST5 XXL}}            & 0.566      & 29                             & distance                  \\
        \textbf{\texttt{EmbeddingGemma}}     & 0.527      & 27                             & distance                  \\
        \textbf{\texttt{Nomic v2}}           & 0.374      & 19                             & distance                  \\
        \textbf{\texttt{Multilang E5 L Ins}} & 0.520      & 27                             & distance                  \\
        \textbf{\texttt{Jina v4}}            & 0.470      & 28                             & distance                  \\
        \textbf{\texttt{Linq Mistral}}       & 0.577      & 32                             & distance                  \\
        \textbf{\texttt{KaLM v2}}            & 0.583      & 32                             & distance                  \\
        \textbf{\texttt{LLaMA Nemotron 8B}}  & 0.449      & 64                             & distance                  \\
        \textbf{\texttt{Qwen3 8B}}           & 0.570      & 35                             & distance                  \\
        \textbf{\texttt{Voyage v3 L}}        & 0.469      & 23                             & distance                  \\
        \textbf{\texttt{OpenAI Text v3 L}}   & 0.560      & 24                             & distance                  \\
        \textbf{\texttt{Gemini 001}}         & 0.554      & 25                             & distance                  \\
        \bottomrule
    \end{tabularx}
\end{table}

\begin{table}[!htb]
    \centering

    \caption{Summary of the best hyperparameters for $k$-NN on NRC-VAD, split with the morphology-aware strategy.}
    \begin{tabularx}{\linewidth}{lXXX}
        \toprule
                                             & $\bm{R^2}$ & \textbf{\texttt{n\_neighbors}} & \textbf{\texttt{weights}} \\
        \midrule
        \textbf{\texttt{ST5 XXL}}            & 0.578      & 27                             & distance                  \\
        \textbf{\texttt{EmbeddingGemma}}     & 0.528      & 32                             & distance                  \\
        \textbf{\texttt{Nomic v2}}           & 0.368      & 27                             & distance                  \\
        \textbf{\texttt{Multilang E5 L Ins}} & 0.524      & 27                             & distance                  \\
        \textbf{\texttt{Jina v4}}            & 0.468      & 34                             & distance                  \\
        \textbf{\texttt{Linq Mistral}}       & 0.587      & 29                             & distance                  \\
        \textbf{\texttt{KaLM v2}}            & 0.594      & 27                             & distance                  \\
        \textbf{\texttt{LLaMA Nemotron 8B}}  & 0.436      & 70                             & distance                  \\
        \textbf{\texttt{Qwen3 8B}}           & 0.574      & 30                             & distance                  \\
        \textbf{\texttt{Voyage v3 L}}        & 0.481      & 21                             & distance                  \\
        \textbf{\texttt{OpenAI Text v3 L}}   & 0.569      & 27                             & distance                  \\
        \textbf{\texttt{Gemini 001}}         & 0.562      & 21                             & distance                  \\
        \bottomrule
    \end{tabularx}
\end{table}

\begin{table}[!htb]
    \centering

    \caption{Summary of the best hyperparameters for $k$-NN on NRC-EIL, split with the semantics-aware strategy.}
    \begin{tabularx}{\linewidth}{lXXX}
        \toprule
                                             & $\bm{R^2}$ & \textbf{\texttt{n\_neighbors}} & \textbf{\texttt{weights}} \\
        \midrule
        \textbf{\texttt{ST5 XXL}}            & 0.464      & 18                             & distance                  \\
        \textbf{\texttt{EmbeddingGemma}}     & 0.484      & 30                             & distance                  \\
        \textbf{\texttt{Nomic v2}}           & 0.291      & 24                             & distance                  \\
        \textbf{\texttt{Multilang E5 L Ins}} & 0.475      & 25                             & distance                  \\
        \textbf{\texttt{Jina v4}}            & 0.367      & 25                             & distance                  \\
        \textbf{\texttt{Linq Mistral}}       & 0.495      & 24                             & distance                  \\
        \textbf{\texttt{KaLM v2}}            & 0.491      & 28                             & distance                  \\
        \textbf{\texttt{LLaMA Nemotron 8B}}  & 0.456      & 22                             & distance                  \\
        \textbf{\texttt{Qwen3 8B}}           & 0.494      & 32                             & distance                  \\
        \textbf{\texttt{Voyage v3 L}}        & 0.381      & 21                             & distance                  \\
        \textbf{\texttt{OpenAI Text v3 L}}   & 0.445      & 16                             & distance                  \\
        \textbf{\texttt{Gemini 001}}         & 0.456      & 25                             & distance                  \\
        \bottomrule
    \end{tabularx}
\end{table}

\begin{table}[!htb]
    \centering

    \caption{Summary of the best hyperparameters for $k$-NN on NRC-EIL, split with the morphology-aware strategy.}
    \begin{tabularx}{\linewidth}{lXXX}
        \toprule
                                             & $\bm{R^2}$ & \textbf{\texttt{n\_neighbors}} & \textbf{\texttt{weights}} \\
        \midrule
        \textbf{\texttt{ST5 XXL}}            & 0.488      & 19                             & distance                  \\
        \textbf{\texttt{EmbeddingGemma}}     & 0.490      & 23                             & distance                  \\
        \textbf{\texttt{Nomic v2}}           & 0.305      & 20                             & distance                  \\
        \textbf{\texttt{Multilang E5 L Ins}} & 0.479      & 24                             & distance                  \\
        \textbf{\texttt{Jina v4}}            & 0.380      & 25                             & distance                  \\
        \textbf{\texttt{Linq Mistral}}       & 0.504      & 23                             & distance                  \\
        \textbf{\texttt{KaLM v2}}            & 0.502      & 31                             & distance                  \\
        \textbf{\texttt{LLaMA Nemotron 8B}}  & 0.457      & 23                             & distance                  \\
        \textbf{\texttt{Qwen3 8B}}           & 0.502      & 27                             & distance                  \\
        \textbf{\texttt{Voyage v3 L}}        & 0.416      & 22                             & distance                  \\
        \textbf{\texttt{OpenAI Text v3 L}}   & 0.471      & 16                             & distance                  \\
        \textbf{\texttt{Gemini 001}}         & 0.470      & 20                             & distance                  \\
        \bottomrule
    \end{tabularx}
\end{table}

\begin{table}[!htb]
    \centering

    \caption{Summary of the best hyperparameters for $k$-NN on GoEmotions.}
    \begin{tabularx}{\linewidth}{lXXX}
        \toprule
                                             & $\bm{F_1}$ & \textbf{\texttt{n\_neighbors}} & \textbf{\texttt{weights}} \\
        \midrule
        \textbf{\texttt{ST5 XXL}}            & 0.410      & 4                              & uniform                   \\
        \textbf{\texttt{EmbeddingGemma}}     & 0.535      & 6                              & uniform                   \\
        \textbf{\texttt{Nomic v2}}           & 0.418      & 4                              & uniform                   \\
        \textbf{\texttt{Multilang E5 L Ins}} & 0.472      & 4                              & uniform                   \\
        \textbf{\texttt{Jina v4}}            & 0.387      & 4                              & uniform                   \\
        \textbf{\texttt{Linq Mistral}}       & 0.487      & 6                              & uniform                   \\
        \textbf{\texttt{KaLM v2}}            & 0.524      & 6                              & uniform                   \\
        \textbf{\texttt{LLaMA Nemotron 8B}}  & 0.418      & 4                              & uniform                   \\
        \textbf{\texttt{Qwen3 8B}}           & 0.512      & 6                              & uniform                   \\
        \textbf{\texttt{Voyage v3 L}}        & 0.398      & 4                              & uniform                   \\
        \textbf{\texttt{OpenAI Text v3 L}}   & 0.440      & 4                              & uniform                   \\
        \textbf{\texttt{Gemini 001}}         & 0.557      & 6                              & uniform                   \\
        \bottomrule
    \end{tabularx}
\end{table}

\begin{table}[!htb]
    \centering

    \caption{Summary of the best hyperparameters for XGB on NRC-VAD, split with the semantics-aware strategy.}
    \begin{tabularx}{\linewidth}{lXXXX}
        \toprule
                                             & $\bm{R^2}$ & \textbf{\texttt{lr}} & \textbf{\texttt{max\_depth}} & \textbf{\texttt{reg\_lambda}} \\
        \midrule
        \textbf{\texttt{ST5 XXL}}            & 0.558      & 0.048                & 7                            & 1.5e-03                       \\
        \textbf{\texttt{EmbeddingGemma}}     & 0.548      & 0.047                & 8                            & 0.368                         \\
        \textbf{\texttt{Nomic v2}}           & 0.391      & 0.064                & 7                            & 5.7e-05                       \\
        \textbf{\texttt{Multilang E5 L Ins}} & 0.538      & 0.041                & 8                            & 1.2e-05                       \\
        \textbf{\texttt{Jina v4}}            & 0.501      & 0.056                & 7                            & 7.2e-06                       \\
        \textbf{\texttt{Linq Mistral}}       & 0.617      & 0.054                & 7                            & 1.6e-08                       \\
        \textbf{\texttt{KaLM v2}}            & 0.623      & 0.047                & 7                            & 5.2e-07                       \\
\bottomrule
    \end{tabularx}
\end{table}

\begin{table}[!htb]
    \centering

    \caption{Summary of the best hyperparameters for XGB on NRC-VAD, split with the morphology-aware strategy.}
    \begin{tabularx}{\linewidth}{lXXXX}
        \toprule
                                             & $\bm{R^2}$ & \textbf{\texttt{lr}} & \textbf{\texttt{max\_depth}} & \textbf{\texttt{reg\_lambda}} \\
        \midrule
        \textbf{\texttt{ST5 XXL}}            & 0.568      & 0.066                & 7                            & 0.010                         \\
        \textbf{\texttt{EmbeddingGemma}}     & 0.547      & 0.040                & 8                            & 0.308                         \\
        \textbf{\texttt{Nomic v2}}           & 0.388      & 0.063                & 7                            & 4.1e-08                       \\
        \textbf{\texttt{Multilang E5 L Ins}} & 0.541      & 0.049                & 8                            & 0.393                         \\
        \textbf{\texttt{Jina v4}}            & 0.502      & 0.055                & 7                            & 0.027                         \\
        \textbf{\texttt{Linq Mistral}}       & 0.627      & 0.058                & 7                            & 2.2e-08                       \\
        \textbf{\texttt{KaLM v2}}            & 0.633      & 0.060                & 7                            & 0.064                         \\
\bottomrule
    \end{tabularx}
\end{table}

\begin{table}[!htb]
    \centering

    \caption{Summary of the best hyperparameters for XGB on NRC-EIL, split with the semantics-aware strategy.}
    \begin{tabularx}{\linewidth}{lXXXX}
        \toprule
                                             & $\bm{R^2}$ & \textbf{\texttt{lr}} & \textbf{\texttt{max\_depth}} & \textbf{\texttt{reg\_lambda}} \\
        \midrule
        \textbf{\texttt{ST5 XXL}}            & 0.432      & 0.052                & 4                            & 5.7e-05                       \\
        \textbf{\texttt{EmbeddingGemma}}     & 0.474      & 0.041                & 4                            & 0.504                         \\
        \textbf{\texttt{Nomic v2}}           & 0.301      & 0.046                & 4                            & 0.712                         \\
        \textbf{\texttt{Multilang E5 L Ins}} & 0.490      & 0.028                & 5                            & 0.710                         \\
        \textbf{\texttt{Jina v4}}            & 0.379      & 0.054                & 3                            & 0.839                         \\
        \textbf{\texttt{Linq Mistral}}       & 0.516      & 0.029                & 4                            & 1.3e-04                       \\
        \textbf{\texttt{KaLM v2}}            & 0.506      & 0.029                & 5                            & 7.0e-06                       \\
\bottomrule
    \end{tabularx}
\end{table}

\begin{table}[!htb]
    \centering

    \caption{Summary of the best hyperparameters for XGB on NRC-EIL, split with the morphology-aware strategy.}
    \begin{tabularx}{\linewidth}{lXXXX}
        \toprule
                                             & $\bm{R^2}$ & \textbf{\texttt{lr}} & \textbf{\texttt{max\_depth}} & \textbf{\texttt{reg\_lambda}} \\
        \midrule
        \textbf{\texttt{ST5 XXL}}            & 0.452      & 0.048                & 4                            & 0.987                         \\
        \textbf{\texttt{EmbeddingGemma}}     & 0.484      & 0.030                & 5                            & 0.631                         \\
        \textbf{\texttt{Nomic v2}}           & 0.306      & 0.048                & 4                            & 0.490                         \\
        \textbf{\texttt{Multilang E5 L Ins}} & 0.489      & 0.026                & 5                            & 0.462                         \\
        \textbf{\texttt{Jina v4}}            & 0.391      & 0.048                & 4                            & 0.962                         \\
        \textbf{\texttt{Linq Mistral}}       & 0.523      & 0.030                & 4                            & 0.985                         \\
        \textbf{\texttt{KaLM v2}}            & 0.518      & 0.030                & 5                            & 1.9e-05                       \\
\bottomrule
    \end{tabularx}
\end{table}

\begin{table}[!htb]
    \centering

    \caption{Summary of the best hyperparameters for XGB on GoEmotions.}
    \begin{tabularx}{\linewidth}{lXXXX}
        \toprule
                                             & $\bm{F_1}$ & \textbf{\texttt{lr}} & \textbf{\texttt{max\_depth}} & \textbf{\texttt{reg\_lambda}} \\
        \midrule
        \textbf{\texttt{ST5 XXL}}            & 0.408      & 0.298                & 3                            & 3.7e-08                       \\
        \textbf{\texttt{EmbeddingGemma}}     & 0.520      & 0.189                & 3                            & 0.996                         \\
        \textbf{\texttt{Nomic v2}}           & 0.425      & 0.299                & 3                            & 0.111                         \\
        \textbf{\texttt{Multilang E5 L Ins}} & 0.464      & 0.214                & 5                            & 0.683                         \\
        \textbf{\texttt{Jina v4}}            & 0.412      & 0.293                & 3                            & 0.982                         \\
        \textbf{\texttt{Linq Mistral}}       & 0.478      & 0.178                & 4                            & 2.3e-06                       \\
        \textbf{\texttt{KaLM v2}}            & 0.521      & 0.141                & 5                            & 0.658                         \\
\bottomrule
    \end{tabularx}
\end{table}

\begin{table}[!htb]
    \centering

    \caption{Summary of the best hyperparameters for MLP on NRC-VAD, split with the semantics-aware strategy.}
    \begin{tabularx}{\linewidth}{lXXXXX}
        \toprule
                                             & $\bm{R^2}$ & \textbf{\texttt{complexity}} & \textbf{\texttt{activation}} & \textbf{\texttt{lr}} & \textbf{\texttt{decay}} \\
        \midrule
        \textbf{\texttt{ST5 XXL}}            & 0.624      & 4                            & relu                         & 1.6e-04              & 8.8e-04                 \\
        \textbf{\texttt{EmbeddingGemma}}     & 0.596      & 4                            & relu                         & 1.4e-04              & 1.2e-03                 \\
        \textbf{\texttt{Nomic v2}}           & 0.462      & 4                            & relu                         & 4.4e-05              & 1.1e-03                 \\
        \textbf{\texttt{Multilang E5 L Ins}} & 0.569      & 4                            & relu                         & 6.0e-05              & 9.8e-04                 \\
        \textbf{\texttt{Jina v4}}            & 0.555      & 4                            & relu                         & 6.8e-05              & 1.9e-03                 \\
        \textbf{\texttt{Linq Mistral}}       & 0.663      & 4                            & relu                         & 4.8e-05              & 1.5e-03                 \\
        \textbf{\texttt{KaLM v2}}            & 0.673      & 4                            & relu                         & 5.8e-05              & 2.1e-03                 \\
\bottomrule
    \end{tabularx}
\end{table}

\begin{table}[!htb]
    \centering

    \caption{Summary of the best hyperparameters for MLP on NRC-VAD, split with the morphology-aware strategy.}
    \begin{tabularx}{\linewidth}{lXXXXX}
        \toprule
                                             & $\bm{R^2}$ & \textbf{\texttt{complexity}} & \textbf{\texttt{activation}} & \textbf{\texttt{lr}} & \textbf{\texttt{decay}} \\
        \midrule
        \textbf{\texttt{ST5 XXL}}            & 0.632      & 4                            & relu                         & 3.2e-05              & 7.9e-04                 \\
        \textbf{\texttt{EmbeddingGemma}}     & 0.594      & 4                            & relu                         & 7.0e-05              & 8.9e-04                 \\
        \textbf{\texttt{Nomic v2}}           & 0.459      & 4                            & relu                         & 2.0e-05              & 1.5e-03                 \\
        \textbf{\texttt{Multilang E5 L Ins}} & 0.574      & 4                            & relu                         & 3.3e-05              & 7.3e-04                 \\
        \textbf{\texttt{Jina v4}}            & 0.558      & 3                            & relu                         & 8.6e-05              & 1.7e-03                 \\
        \textbf{\texttt{Linq Mistral}}       & 0.676      & 4                            & relu                         & 4.7e-05              & 1.5e-03                 \\
        \textbf{\texttt{KaLM v2}}            & 0.681      & 4                            & relu                         & 4.8e-05              & 2.0e-03                 \\
\bottomrule
    \end{tabularx}
\end{table}

\begin{table}[!htb]
    \centering

    \caption{Summary of the best hyperparameters for MLP on NRC-EIL, split with the semantics-aware strategy.}
    \begin{tabularx}{\linewidth}{lXXXXX}
        \toprule
                                             & $\bm{R^2}$ & \textbf{\texttt{complexity}} & \textbf{\texttt{activation}} & \textbf{\texttt{lr}} & \textbf{\texttt{decay}} \\
        \midrule
        \textbf{\texttt{ST5 XXL}}            & 0.506      & 8                            & relu                         & 2.0e-05              & 2.0e-05                 \\
        \textbf{\texttt{EmbeddingGemma}}     & 0.523      & 8                            & relu                         & 3.4e-05              & 1.6e-05                 \\
        \textbf{\texttt{Nomic v2}}           & 0.374      & 4                            & relu                         & 5.2e-05              & 1.3e-05                 \\
        \textbf{\texttt{Multilang E5 L Ins}} & 0.504      & 8                            & relu                         & 3.7e-05              & 1.7e-05                 \\
        \textbf{\texttt{Jina v4}}            & 0.431      & 1                            & logistic                     & 2.9e-05              & 1.3e-05                 \\
        \textbf{\texttt{Linq Mistral}}       & 0.539      & 8                            & relu                         & 1.2e-05              & 1.0e-05                 \\
        \textbf{\texttt{KaLM v2}}            & 0.541      & 8                            & relu                         & 1.0e-05              & 2.5e-05                 \\
\bottomrule
    \end{tabularx}
\end{table}

\begin{table}[!htb]
    \centering

    \caption{Summary of the best hyperparameters for MLP on NRC-EIL, split with the morphology-aware strategy.}
    \begin{tabularx}{\linewidth}{lXXXXX}
        \toprule
                                             & $\bm{R^2}$ & \textbf{\texttt{complexity}} & \textbf{\texttt{activation}} & \textbf{\texttt{lr}} & \textbf{\texttt{decay}} \\
        \midrule
        \textbf{\texttt{ST5 XXL}}            & 0.522      & 8                            & relu                         & 7.5e-05              & 1.5e-05                 \\
        \textbf{\texttt{EmbeddingGemma}}     & 0.524      & 8                            & relu                         & 6.6e-05              & 1.0e-05                 \\
        \textbf{\texttt{Nomic v2}}           & 0.378      & 4                            & relu                         & 4.6e-05              & 4.6e-05                 \\
        \textbf{\texttt{Multilang E5 L Ins}} & 0.503      & 4                            & relu                         & 1.2e-04              & 2.2e-05                 \\
        \textbf{\texttt{Jina v4}}            & 0.436      & 1                            & logistic                     & 3.1e-05              & 1.0e-05                 \\
        \textbf{\texttt{Linq Mistral}}       & 0.544      & 8                            & relu                         & 3.4e-05              & 5.8e-05                 \\
        \textbf{\texttt{KaLM v2}}            & 0.544      & 7                            & relu                         & 7.5e-05              & 6.6e-05                 \\
\bottomrule
    \end{tabularx}
\end{table}

\begin{table}[!htb]
    \centering

    \caption{Summary of the best hyperparameters for MLP on GoEmotions.}
    \label{tab:opt_last}
    \begin{tabularx}{\linewidth}{lXXXXX}
        \toprule
                                             & $\bm{F_1}$ & \textbf{\texttt{complexity}} & \textbf{\texttt{activation}} & \textbf{\texttt{lr}} & \textbf{\texttt{decay}} \\
        \midrule
        \textbf{\texttt{ST5 XXL}}            & 0.501      & 7                            & tanh                         & 9.7e-04              & 1.0e-05                 \\
        \textbf{\texttt{EmbeddingGemma}}     & 0.571      & 7                            & tanh                         & 2.5e-03              & 2.9e-05                 \\
        \textbf{\texttt{Nomic v2}}           & 0.520      & 10                           & tanh                         & 6.7e-04              & 2.2e-05                 \\
        \textbf{\texttt{Multilang E5 L Ins}} & 0.531      & 8                            & tanh                         & 1.9e-03              & 2.2e-05                 \\
        \textbf{\texttt{Jina v4}}            & 0.500      & 3                            & tanh                         & 1.9e-03              & 1.2e-05                 \\
        \textbf{\texttt{Linq Mistral}}       & 0.551      & 3                            & tanh                         & 1.3e-03              & 1.0e-05                 \\
        \textbf{\texttt{KaLM v2}}            & 0.571      & 4                            & tanh                         & 8.4e-04              & 3.1e-05                 \\
\bottomrule
    \end{tabularx}
\end{table}

\begin{table}[!htb]
    \centering

    \caption{
        Full summary of the 5-fold cross-validation scores for LR and $k$-NN on NRC-VAD.
        For each encoder, top and bottom rows refer to data splits based on semantic and morphological leakage prevention, respectively.
    }
    \label{tab:NRC-VAD_Full_Metrics_1}
    \begin{tabularx}{\linewidth}{X*{6}{c}}
                                                              & \multicolumn{3}{c}{\textbf{LR}} & \multicolumn{3}{c}{$\bm{k}$\textbf{-NN}}                                                                              \\
                                                              & MSE                             & $R^2$                                    & $\rho_{c}$       & MSE               & $R^2$            & $\rho_{c}$       \\
        \cmidrule(l{1.75pt}r{1.75pt}){2-4}\cmidrule(l{1.75pt}r{1.75pt}){5-7}
        \multirow{2}{*}{\textbf{\texttt{KaLM v2}}}            & $0.070_{\pm.003}$               & $.631_{\pm.014}$                         & $.772_{\pm.009}$ & $0.079_{\pm.004}$ & $.583_{\pm.014}$ & $.720_{\pm.010}$ \\
                                                              & $0.071_{\pm.001}$               & $.640_{\pm.008}$                         & $.780_{\pm.005}$ & $0.080_{\pm.001}$ & $.594_{\pm.007}$ & $.731_{\pm.004}$ \\
        \cmidrule(l{1.75pt}r{1.75pt}){2-4}\cmidrule(l{1.75pt}r{1.75pt}){5-7}
        \multirow{2}{*}{\textbf{\texttt{Linq Mistral}}}       & $0.071_{\pm.003}$               & $.626_{\pm.014}$                         & $.768_{\pm.009}$ & $0.080_{\pm.004}$ & $.577_{\pm.013}$ & $.717_{\pm.011}$ \\
                                                              & $0.072_{\pm.001}$               & $.637_{\pm.009}$                         & $.778_{\pm.006}$ & $0.082_{\pm.001}$ & $.587_{\pm.007}$ & $.725_{\pm.004}$ \\
        \cmidrule(l{1.75pt}r{1.75pt}){2-4}\cmidrule(l{1.75pt}r{1.75pt}){5-7}
        \multirow{2}{*}{\textbf{\texttt{OpenAI Text v3 L}}}   & $0.077_{\pm.004}$               & $.597_{\pm.010}$                         & $.745_{\pm.008}$ & $0.084_{\pm.004}$ & $.560_{\pm.010}$ & $.682_{\pm.008}$ \\
\end{tabularx}
\end{table}

\begin{table}[!htb]
    \centering

    \caption{
        Full summary of the 5-fold cross-validation scores for XGB and MLP on NRC-VAD.
        For each encoder, top and bottom rows refer to data splits based on semantic and morphological leakage prevention, respectively.
    }
    \label{tab:NRC-VAD_Full_Metrics_2}
    \begin{tabularx}{\linewidth}{X*{6}{c}}
                                                              & \multicolumn{3}{c}{\textbf{XGB}} & \multicolumn{3}{c}{\textbf{MLP}}                                                                              \\
                                                              & MSE                              & $R^2$                            & $\rho_{c}$       & MSE               & $R^2$            & $\rho_{c}$       \\
        \cmidrule(l{1.75pt}r{1.75pt}){2-4}\cmidrule(l{1.75pt}r{1.75pt}){5-7}
        \multirow{2}{*}{\textbf{\texttt{KaLM v2}}}            & $0.072_{\pm.003}$                & $.623_{\pm.014}$                 & $.756_{\pm.009}$ & $0.062_{\pm.003}$ & $.672_{\pm.013}$ & $.799_{\pm.009}$ \\
                                                              & $0.073_{\pm.001}$                & $.633_{\pm.007}$                 & $.767_{\pm.004}$ & $0.063_{\pm.001}$ & $.681_{\pm.007}$ & $.805_{\pm.004}$ \\
        \cmidrule(l{1.75pt}r{1.75pt}){2-4}\cmidrule(l{1.75pt}r{1.75pt}){5-7}
        \multirow{2}{*}{\textbf{\texttt{Linq Mistral}}}       & $0.073_{\pm.003}$                & $.617_{\pm.014}$                 & $.753_{\pm.010}$ & $0.064_{\pm.003}$ & $.662_{\pm.015}$ & $.791_{\pm.011}$ \\
                                                              & $0.074_{\pm.001}$                & $.627_{\pm.007}$                 & $.761_{\pm.004}$ & $0.065_{\pm.001}$ & $.675_{\pm.009}$ & $.803_{\pm.005}$ \\
        \cmidrule(l{1.75pt}r{1.75pt}){2-4}\cmidrule(l{1.75pt}r{1.75pt}){5-7}
        \multirow{2}{*}{\textbf{\texttt{OpenAI Text v3 L}}}   & $0.082_{\pm.004}$                & $.570_{\pm.013}$                 & $.712_{\pm.010}$ & $0.066_{\pm.003}$ & $.651_{\pm.009}$ & $.783_{\pm.006}$ \\
                                                              & $0.084_{\pm.002}$                & $.579_{\pm.006}$                 & $.720_{\pm.004}$ & $0.068_{\pm.001}$ & $.659_{\pm.006}$ & $.790_{\pm.004}$ \\
\end{tabularx}
\end{table}

\begin{table}[!htb]
    \centering

    \caption{
        Full summary of the regression metrics at test time for NRC-VAD, sorted by $R^{2}$ score of the MLP model in descending order.
        For each encoder, top and bottom rows refer to data splits based on semantic and morphological leakage prevention, respectively.
    }
    \label{tab:NRC-VAD_Full}
    \begin{tabularx}{\linewidth}{X*{12}{c}}
                                                              & \multicolumn{3}{c}{\textbf{LR}} & \multicolumn{3}{c}{$\bm{k}$\textbf{-NN}} & \multicolumn{3}{c}{\textbf{XGB}} & \multicolumn{3}{c}{\textbf{MLP}}                                                                                                                                                                                                                                                                         \\
                                                              & MSE                             & $R^2$                                    & $\rho_{c}$                       & MSE                              & $R^2$                          & $\rho_{c}$                     & MSE                            & $R^2$                          & $\rho_{c}$                     & MSE                            & $R^2$                          & $\rho_{c}$                     \\
        \cmidrule(l{1.75pt}r{1.75pt}){2-4}\cmidrule(l{1.75pt}r{1.75pt}){5-7}\cmidrule(l{1.75pt}r{1.75pt}){8-10}\cmidrule(l{1.75pt}r{1.75pt}){11-13}
        \multirow{2}{*}{\textbf{\texttt{KaLM v2}}}            & $\mathbf{.066}$                 & $\mathbf{.637}$                          & $\mathbf{.790}$                  & $\mathbf{.075}$                  & $\mathbf{.591}$                & $\underline{\underline{.746}}$ & $\mathbf{.068}$                & $\mathbf{.630}$                & $\mathbf{.777}$                & $\mathbf{.059}$                & $\mathbf{.677}$                & $\mathbf{.811}$                \\
\end{tabularx}
\end{table}

\begin{table}[!htb]
    \centering

    \caption{
        Full summary of the 5-fold cross-validation scores for LR and $k$-NN on NRC-EIL.
        For each encoder, top and bottom rows refer to data splits based on semantic and morphological leakage prevention, respectively.
    }
    \label{tab:NRC-EIL_Full_Metrics_1}
    \begin{tabularx}{\linewidth}{X*{6}{c}}
                                                              & \multicolumn{3}{c}{\textbf{LR}} & \multicolumn{3}{c}{$\bm{k}$\textbf{-NN}}                                                                              \\
                                                              & MSE                             & $R^2$                                    & $\rho_{c}$       & MSE               & $R^2$            & $\rho_{c}$       \\
        \cmidrule(l{1.75pt}r{1.75pt}){2-4}\cmidrule(l{1.75pt}r{1.75pt}){5-7}
        \multirow{2}{*}{\textbf{\texttt{KaLM v2}}}            & $0.023_{\pm.000}$               & $.505_{\pm.012}$                         & $.658_{\pm.010}$ & $0.024_{\pm.000}$ & $.491_{\pm.010}$ & $.630_{\pm.007}$ \\
                                                              & $0.022_{\pm.000}$               & $.513_{\pm.010}$                         & $.667_{\pm.006}$ & $0.023_{\pm.000}$ & $.502_{\pm.011}$ & $.638_{\pm.007}$ \\
        \cmidrule(l{1.75pt}r{1.75pt}){2-4}\cmidrule(l{1.75pt}r{1.75pt}){5-7}
        \multirow{2}{*}{\textbf{\texttt{Linq Mistral}}}       & $0.023_{\pm.000}$               & $.515_{\pm.012}$                         & $.666_{\pm.009}$ & $0.023_{\pm.000}$ & $.495_{\pm.010}$ & $.639_{\pm.004}$ \\
                                                              & $0.022_{\pm.000}$               & $.524_{\pm.006}$                         & $.675_{\pm.005}$ & $0.022_{\pm.000}$ & $.504_{\pm.007}$ & $.651_{\pm.005}$ \\
        \cmidrule(l{1.75pt}r{1.75pt}){2-4}\cmidrule(l{1.75pt}r{1.75pt}){5-7}
        \multirow{2}{*}{\textbf{\texttt{Qwen3 8B}}}           & $0.023_{\pm.000}$               & $.500_{\pm.010}$                         & $.653_{\pm.008}$ & $0.024_{\pm.000}$ & $.494_{\pm.013}$ & $.635_{\pm.008}$ \\
\end{tabularx}
\end{table}

\begin{table}[!htb]
    \centering

    \caption{
        Full summary of the 5-fold cross-validation scores for XGB and MLP on NRC-EIL.
        For each encoder, top and bottom rows refer to data splits based on semantic and morphological leakage prevention, respectively.
    }
    \label{tab:NRC-EIL_Full_Metrics_2}
    \begin{tabularx}{\linewidth}{X*{6}{c}}
                                                              & \multicolumn{3}{c}{\textbf{XGB}} & \multicolumn{3}{c}{\textbf{MLP}}                                                                              \\
                                                              & MSE                              & $R^2$                            & $\rho_{c}$       & MSE               & $R^2$            & $\rho_{c}$       \\
        \cmidrule(l{1.75pt}r{1.75pt}){2-4}\cmidrule(l{1.75pt}r{1.75pt}){5-7}
        \multirow{2}{*}{\textbf{\texttt{KaLM v2}}}            & $0.023_{\pm.001}$                & $.506_{\pm.010}$                 & $.646_{\pm.009}$ & $0.021_{\pm.000}$ & $.540_{\pm.015}$ & $.694_{\pm.015}$ \\
                                                              & $0.022_{\pm.001}$                & $.518_{\pm.011}$                 & $.657_{\pm.007}$ & $0.021_{\pm.000}$ & $.543_{\pm.012}$ & $.698_{\pm.008}$ \\
        \cmidrule(l{1.75pt}r{1.75pt}){2-4}\cmidrule(l{1.75pt}r{1.75pt}){5-7}
        \multirow{2}{*}{\textbf{\texttt{Linq Mistral}}}       & $0.022_{\pm.000}$                & $.516_{\pm.010}$                 & $.657_{\pm.008}$ & $0.022_{\pm.001}$ & $.536_{\pm.017}$ & $.693_{\pm.009}$ \\
                                                              & $0.022_{\pm.000}$                & $.523_{\pm.009}$                 & $.663_{\pm.008}$ & $0.021_{\pm.000}$ & $.542_{\pm.008}$ & $.696_{\pm.006}$ \\
        \cmidrule(l{1.75pt}r{1.75pt}){2-4}\cmidrule(l{1.75pt}r{1.75pt}){5-7}
        \multirow{2}{*}{\textbf{\texttt{Qwen3 8B}}}           & $0.023_{\pm.001}$                & $.501_{\pm.015}$                 & $.646_{\pm.012}$ & $0.022_{\pm.001}$ & $.526_{\pm.015}$ & $.681_{\pm.011}$ \\
                                                              & $0.022_{\pm.000}$                & $.508_{\pm.004}$                 & $.647_{\pm.003}$ & $0.021_{\pm.000}$ & $.527_{\pm.009}$ & $.680_{\pm.006}$ \\
\end{tabularx}
\end{table}

\begin{table}[!htb]
    \centering

    \caption{
        Full summary of the regression metrics at test time for NRC-EIL, sorted by $R^{2}$ score of the MLP model in descending order.
        For each encoder, top and bottom rows refer to data splits based on semantic and morphological leakage prevention, respectively.
    }
    \label{tab:NRC-EIL_Full}
    \begin{tabularx}{\linewidth}{X*{12}{c}}
                                                              & \multicolumn{3}{c}{\textbf{LR}} & \multicolumn{3}{c}{$\bm{k}$\textbf{-NN}} & \multicolumn{3}{c}{\textbf{XGB}} & \multicolumn{3}{c}{\textbf{MLP}}                                                                                                                                                                                                                                                             \\
                                                              & MSE                             & $R^2$                                    & $\rho_{c}$                       & MSE                              & $R^2$                          & $\rho_{c}$         & MSE                            & $R^2$                          & $\rho_{c}$                     & MSE                            & $R^2$                          & $\rho_{c}$                     \\
        \cmidrule(l{1.75pt}r{1.75pt}){2-4}\cmidrule(l{1.75pt}r{1.75pt}){5-7}\cmidrule(l{1.75pt}r{1.75pt}){8-10}\cmidrule(l{1.75pt}r{1.75pt}){11-13}
        \multirow{2}{*}{\textbf{\texttt{KaLM v2}}}            & $\underline{\underline{.023}}$  & $\underline{\underline{.516}}$           & $\underline{\underline{.705}}$   & $\mathbf{.024}$                  & $\underline{\underline{.500}}$ & $\underline{.685}$ & $\underline{\underline{.023}}$ & $\underline{\underline{.515}}$ & $\underline{\underline{.697}}$ & $\mathbf{.022}$                & $\mathbf{.540}$                & $\mathbf{.730}$                \\
\end{tabularx}
\end{table}

\begin{table}[!htb]
    \centering

    \caption{
        Full summary of the 5-fold cross-validation scores for LR and $k$-NN on GoEmotions.
        For each encoder, its rows refer to macro, weighted, and micro averages, respectively.
    }
    \label{tab:GoEmotions_Full_Metrics_1}
    \begin{tabularx}{\linewidth}{X*{6}{c}}
                                                              & \multicolumn{3}{c}{\textbf{LR}} & \multicolumn{3}{c}{$\bm{k}$\textbf{-NN}}                                                                             \\
                                                              & \texttt{p}                      & \texttt{r}                               & $F_{1}$          & \texttt{p}       & \texttt{r}       & $F_{1}$          \\
        \cmidrule(l{1.75pt}r{1.75pt}){2-4}\cmidrule(l{1.75pt}r{1.75pt}){5-7}
        \multirow{3}{*}{\textbf{\texttt{Gemini 001}}}         & $.698_{\pm.004}$                & $.495_{\pm.005}$                         & $.573_{\pm.005}$ & $.596_{\pm.008}$ & $.529_{\pm.006}$ & $.557_{\pm.005}$ \\
                                                              & $.731_{\pm.004}$                & $.608_{\pm.004}$                         & $.660_{\pm.003}$ & $.646_{\pm.002}$ & $.630_{\pm.004}$ & $.636_{\pm.003}$ \\
                                                              & $.739_{\pm.004}$                & $.608_{\pm.004}$                         & $.667_{\pm.003}$ & $.652_{\pm.002}$ & $.630_{\pm.004}$ & $.640_{\pm.003}$ \\
        \cmidrule(l{1.75pt}r{1.75pt}){2-4}\cmidrule(l{1.75pt}r{1.75pt}){5-7}
        \multirow{3}{*}{\textbf{\texttt{EmbeddingGemma}}}     & $.679_{\pm.006}$                & $.483_{\pm.007}$                         & $.559_{\pm.006}$ & $.586_{\pm.009}$ & $.501_{\pm.004}$ & $.535_{\pm.004}$ \\
                                                              & $.724_{\pm.001}$                & $.590_{\pm.004}$                         & $.647_{\pm.002}$ & $.629_{\pm.002}$ & $.606_{\pm.003}$ & $.614_{\pm.002}$ \\
                                                              & $.734_{\pm.001}$                & $.590_{\pm.004}$                         & $.654_{\pm.002}$ & $.635_{\pm.002}$ & $.606_{\pm.003}$ & $.620_{\pm.002}$ \\
\end{tabularx}
\end{table}

\begin{table}[!htb]
    \centering

    \caption{
        Full summary of the 5-fold cross-validation scores for XGB and MLP on GoEmotions.
        For each encoder, its rows refer to macro, weighted, and micro averages, respectively.
    }
    \label{tab:GoEmotions_Full_Metrics_2}
    \begin{tabularx}{\linewidth}{X*{6}{c}}
                                                              & \multicolumn{3}{c}{\textbf{XGB}} & \multicolumn{3}{c}{\textbf{MLP}}                                                                             \\
                                                              & \texttt{p}                       & \texttt{r}                       & $F_{1}$          & \texttt{p}       & \texttt{r}       & $F_{1}$          \\
        \cmidrule(l{1.75pt}r{1.75pt}){2-4}\cmidrule(l{1.75pt}r{1.75pt}){5-7}
        \multirow{3}{*}{\textbf{\texttt{Gemini 001}}}         & $.656_{\pm.009}$                 & $.464_{\pm.005}$                 & $.538_{\pm.005}$ & $.688_{\pm.009}$ & $.508_{\pm.006}$ & $.577_{\pm.006}$ \\
                                                              & $.712_{\pm.004}$                 & $.577_{\pm.003}$                 & $.634_{\pm.002}$ & $.721_{\pm.005}$ & $.622_{\pm.005}$ & $.665_{\pm.005}$ \\
                                                              & $.722_{\pm.004}$                 & $.577_{\pm.003}$                 & $.641_{\pm.002}$ & $.728_{\pm.004}$ & $.622_{\pm.005}$ & $.671_{\pm.004}$ \\
        \cmidrule(l{1.75pt}r{1.75pt}){2-4}\cmidrule(l{1.75pt}r{1.75pt}){5-7}
        \multirow{3}{*}{\textbf{\texttt{EmbeddingGemma}}}     & $.672_{\pm.004}$                 & $.432_{\pm.006}$                 & $.520_{\pm.005}$ & $.671_{\pm.014}$ & $.493_{\pm.010}$ & $.559_{\pm.009}$ \\
                                                              & $.716_{\pm.002}$                 & $.543_{\pm.002}$                 & $.613_{\pm.002}$ & $.715_{\pm.003}$ & $.608_{\pm.013}$ & $.652_{\pm.008}$ \\
                                                              & $.729_{\pm.002}$                 & $.543_{\pm.002}$                 & $.622_{\pm.002}$ & $.720_{\pm.005}$ & $.608_{\pm.013}$ & $.659_{\pm.007}$ \\
\end{tabularx}
\end{table}

\begin{table}[!htb]
    \centering

    \caption{
        Full summary of the classification metrics at test time for GoEmotions, sorted by $F_{1}$ score of the MLP model in descending order.
        For each encoder, its rows refer to macro, weighted, and micro averages, respectively.
    }
    \label{tab:GoEmotions_Full}
    \begin{tabular}{l*{12}{c}}
                                                              & \multicolumn{3}{c}{\textbf{LR}} & \multicolumn{3}{c}{$\bm{k}$\textbf{-NN}} & \multicolumn{3}{c}{\textbf{XGB}} & \multicolumn{3}{c}{\textbf{MLP}}                                                                                                                                                                                                                                           \\
                                                              & \texttt{p}                      & \texttt{r}                               & $F_{1}$                          & \texttt{p}                       & \texttt{r}      & $F_{1}$         & \texttt{p}                     & \texttt{r}                     & $F_{1}$                        & \texttt{p}                     & \texttt{r}                     & $F_{1}$                        \\
        \cmidrule(l{1.75pt}r{1.75pt}){2-4}\cmidrule(l{1.75pt}r{1.75pt}){5-7}\cmidrule(l{1.75pt}r{1.75pt}){8-10}\cmidrule(l{1.75pt}r{1.75pt}){11-13}
        \multirow{3}{*}{\textbf{\texttt{Gemini 001}}}         & $\mathbf{.716}$                 & $\mathbf{.517}$                          & $\mathbf{.594}$                  & $\mathbf{.624}$                  & $\mathbf{.543}$ & $\mathbf{.575}$ & $\underline{\underline{.674}}$ & $\mathbf{.469}$                & $\mathbf{.546}$                & $\mathbf{.713}$                & $\mathbf{.529}$                & $\mathbf{.600}$                \\
                                                              & $\mathbf{.735}$                 & $\mathbf{.612}$                          & $\mathbf{.664}$                  & $\mathbf{.649}$                  & $\mathbf{.631}$ & $\mathbf{.637}$ & $\underline{\underline{.714}}$ & $\mathbf{.579}$                & $\mathbf{.636}$                & $\underline{\underline{.719}}$ & $\mathbf{.623}$                & $\mathbf{.663}$                \\
\end{tabular}
\end{table}

\endgroup

\end{document}